%% file: main.tex
\documentclass[acmtog]{acmart}

\usepackage{amsmath}
\usepackage{multirow}

\AtBeginDocument{%
  \providecommand\BibTeX{{%
    \normalfont B\kern-0.5em{\scshape i\kern-0.25em b}\kern-0.8em\TeX}}}

\definecolor{Purple}{rgb}{.8, 0, .8}
\definecolor{Red}{rgb}{1., 0, 0}
\definecolor{Blue}{rgb}{0, 0, 1.}
\definecolor{Green}{rgb}{0., .6, 0.}
\definecolor{Yellow}{rgb}{.9, .7, 0.}
\definecolor{Orange}{rgb}{1, 0.5, 0.}

\begin{document}

\title{Transposer: Universal Texture Synthesis Using Feature Maps as Transposed Convolution Filter}

\author{Guilin Liu}
\affiliation{%
  \institution{NVIDIA}
}
\author{Rohan Taori}
\affiliation{%
  \institution{NVIDIA, UC Berkeley}
}
\author{Ting-Chun Wang}
\affiliation{%
  \institution{NVIDIA}
}
\author{Zhiding Yu}
\author{Shiqiu Liu}
\affiliation{%
  \institution{NVIDIA}
}
\author{Fitsum A. Reda}
\author{Karan Sapra}
\affiliation{%
  \institution{NVIDIA}
}
\author{Andrew Tao}
\author{Bryan Catanzaro}
\affiliation{%
  \institution{NVIDIA}
}

% \author{Ting-Chun Wang}
% % \email{webmaster@marysville-ohio.com}
% \affiliation{%

\renewcommand{\shortauthors}{Liu, et al.}

\begin{abstract}
Conventional CNNs for texture synthesis consist of a sequence of (de)-convolution and up/down-sampling layers, where each layer operates locally and lacks the ability to capture the long-term structural dependency required by texture synthesis. Thus, they often simply enlarge the input texture, rather than perform reasonable synthesis. As a compromise, many recent methods sacrifice generalizability by training and testing on the same single (or fixed set of) texture image(s), resulting in huge re-training time costs for unseen images. In this work, based on the discovery that the assembling/stitching operation in traditional texture synthesis is analogous to a transposed convolution operation, we propose a novel way of using transposed convolution operation. Specifically, we directly treat \textit{the whole encoded feature map of the input texture} as \textit{transposed convolution filters} and \textit{the features' self-similarity map}, which captures the auto-correlation information, as \textit{input to the transposed convolution}. Such a design allows our framework, once trained, to be generalizable to perform synthesis of unseen textures with a single forward pass in nearly real-time. Our method achieves state-of-the-art texture synthesis quality based on various metrics. While self-similarity helps preserve the input textures' regular structural patterns, our framework can also take random noise maps for irregular input textures instead of self-similarity maps as transposed convolution inputs. It allows to get more diverse results as well as generate arbitrarily large texture outputs by directly sampling large noise maps in a single pass as well.

%Our framework can also take random noise maps as transposed convolution inputs to get diverse results and directly synthesize arbitrarily large outputs.
\end{abstract}






\begin{teaserfigure}
\centering
\addtolength{\tabcolsep}{-3pt}
\small
\scalebox{0.96}{
% \addtolength{\tabcolsep}{-4pt}   
% \begin{tabular}{c|cccccccc}
\begin{tabular}{c|cccccccc}
% \setlength{\tabcolsep}{1em}
% \multicolumn{1}{c}{} & \multicolumn{6}{c}{} \\
% \setlength{\tabcolsep}{1em}
\multicolumn{1}{c|}{input} & \multicolumn{8}{c}{ 4$\times$4 times larger synthesis using Transposer (ours)} \\
\hline
\includegraphics[width=0.056\textwidth]{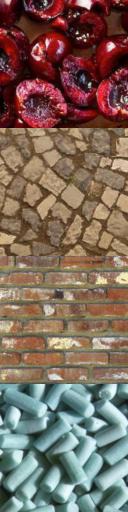} &
\multicolumn{2}{c}{\includegraphics[width=0.224\textwidth]{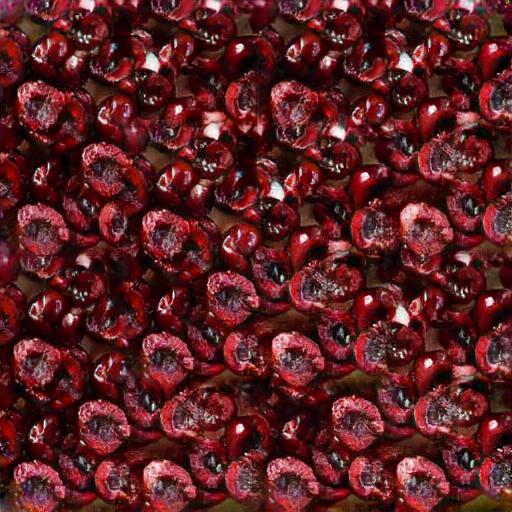}} &
\multicolumn{2}{c}{\includegraphics[width=0.224\textwidth]{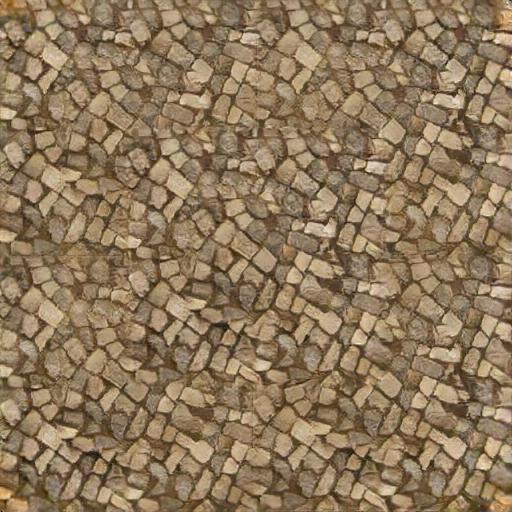}} &
% \multicolumn{2}{c}{\includegraphics[width=0.32\textwidth]{teaser512/213_c1l5r120_output_output.jpg}} & 
\multicolumn{2}{c}{\includegraphics[width=0.224\textwidth]{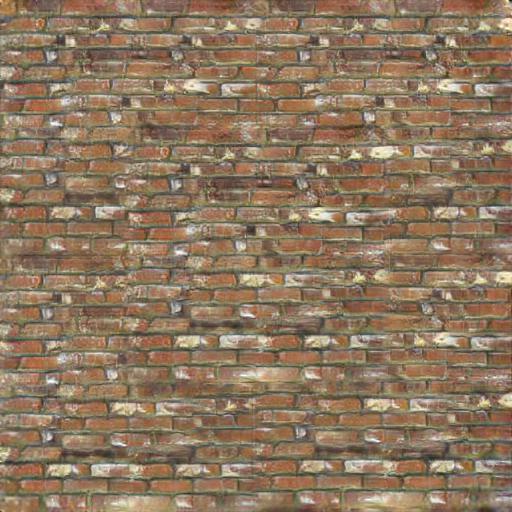}} & 
\multicolumn{2}{c}{\includegraphics[width=0.224\textwidth]{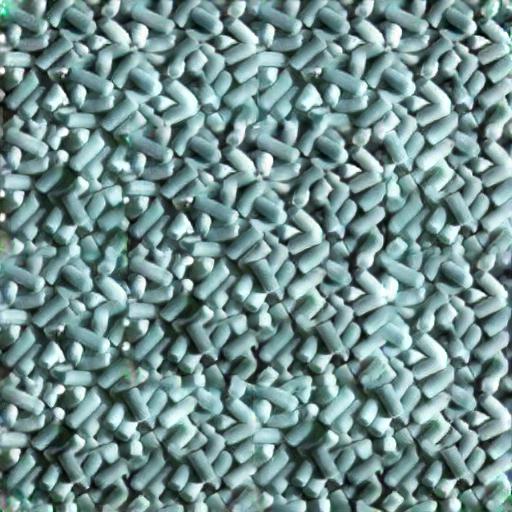}}
\\
\hline
\hline
input & Transposer(ours)  & Self-Tuning & pix2pixHD & Text.Mixer &  SinGAN &  Non-stat. & WCT & DeepTexture\\
\hline
\includegraphics[width=0.056\textwidth]{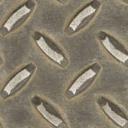} &
\includegraphics[width=0.112\textwidth]{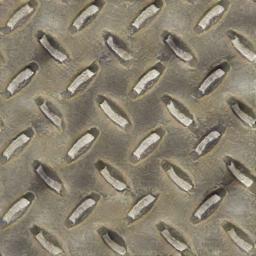} &
\includegraphics[width=0.112\textwidth]{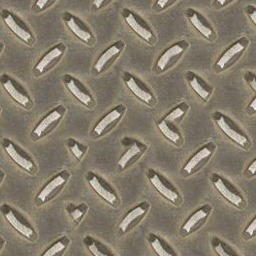} &
\includegraphics[width=0.112\textwidth]{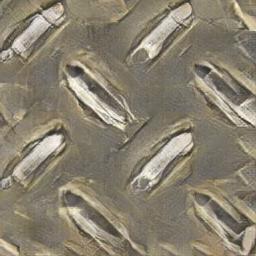} &
\includegraphics[width=0.112\textwidth]{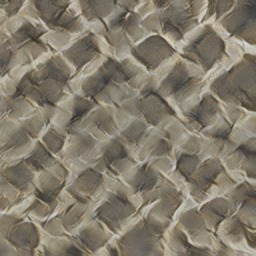} & 
\includegraphics[width=0.112\textwidth]{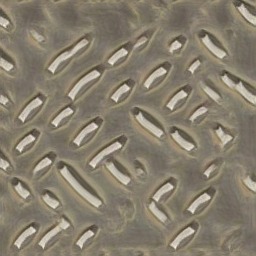} &
\includegraphics[width=0.112\textwidth]{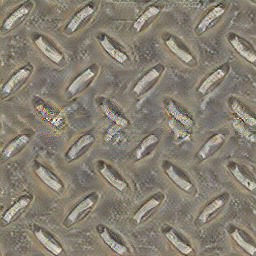} &
\includegraphics[width=0.112\textwidth]{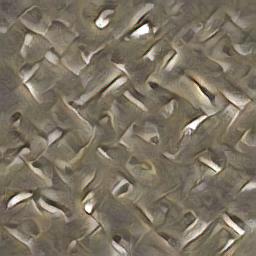} &
\includegraphics[width=0.112\textwidth]{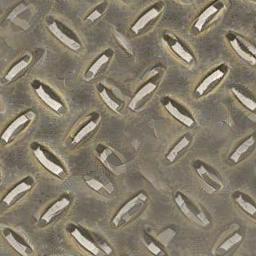} \\
\includegraphics[width=0.056\textwidth]{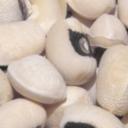} &
\includegraphics[width=0.112\textwidth]{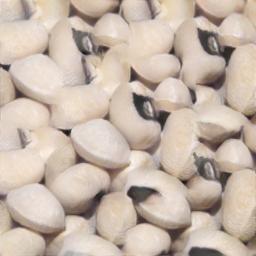} &
\includegraphics[width=0.112\textwidth]{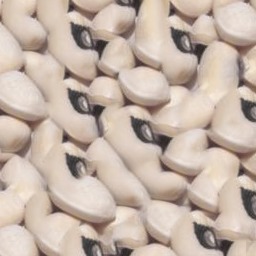} &
\includegraphics[width=0.112\textwidth]{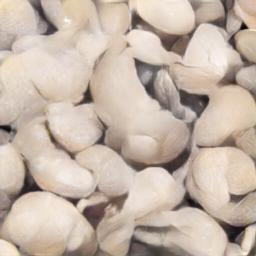} &
\includegraphics[width=0.112\textwidth]{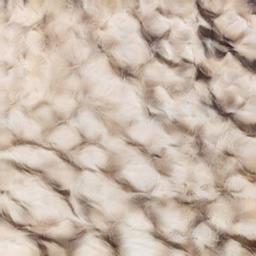} &
\includegraphics[width=0.112\textwidth]{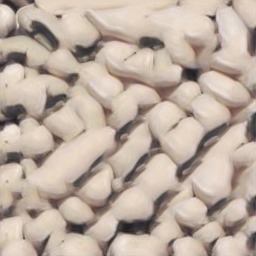} &
\includegraphics[width=0.112\textwidth]{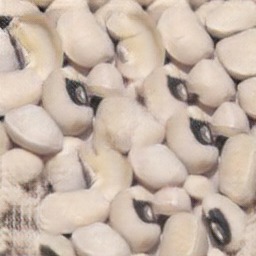} &
\includegraphics[width=0.112\textwidth]{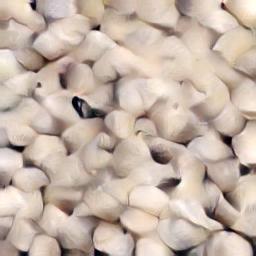} &
\includegraphics[width=0.112\textwidth]{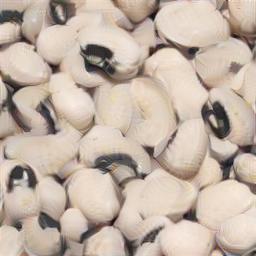} \\
\end{tabular}
% \addtolength{\tabcolsep}{-4pt}   
}
\caption{The first row shows our 4$\times$4 times larger texture synthesis results. The remaining two rows show the texture synthesis results using different approaches. Transposer (ours) represents our method, which is generalizable and can perform texture synthesis on unseen texture images with a single network forward pass in tens or hundreds of milliseconds. Self-Tuning~\cite{kaspar2015self} sometimes fails to fully preserve the regular structure and needs hundreds of seconds to solve the objective function. pix2pixHD~\cite{wang2018high} simply enlarges the input rather than perform synthesis. Texture Mixer~\cite{yu2019texture} cannot handle inputs with structures. sinGAN~\cite{shaham2019singan} and Non-stat.~\cite{zhou2018non} need to take tens of minutes or several hours retrain their models for each input texture. WCT~\cite{yijun2017universal}, the style transfer based method can't preserve structure patterns. DeepTexture~\cite{gatys2015texture} is optimization based method and needs tens of minutes.}
\label{fig:teaser_comp} 
\end{teaserfigure}

\maketitle

\input{sec_intro}

\input{sec_relatedwork}

\input{sec_approach}

\input{sec_comp}

\input{sec_ablation}

\input{sec_discuss}

\begin{acks}
We would like to thanks Brandon Rowllet, Sifei Liu, Aysegul Dundar, Kevin Shih, Rafael Valle and Robert Pottorff for valuable discussions and proof-reading. 
\end{acks}

\bibliographystyle{ACM-Reference-Format}
\bibliography{egbib}

\onecolumn
\appendix

\section{Framework Detail}

\subsection{Self-Similarity Computing \& Transposed Convolution}

The reviewers are welcome to check the attached animation video showing how a self-similarity map is computed and how the transposed convolution operation is performed.
% We provide an animation video of how we compute self-similarity map and how the transposed convolution operation is done. Please check the animation video for more details.

\subsection{Implementation Details for Computing Self-similarity Map}
Computing self-similarity map can be efficiently implemented with the help of standard convolution operations. The formula for computing self-similarity map can be relaxed as the following:
\begin{equation}
    \mathbf{s}(p,q)  = -\frac{\sum_{m, n, c} \|\mathbb{F}^{c}(m, n) - \mathbb{F}^{c}(m-p, n-q)\|^2}{\sum_{m, n, c} \|\mathbb{F}^{c}(m, n)\|^2}
     = -\frac{\sum_{m, n, c} \|\mathbb{F}^{c}(m, n)\|^2 - 2*\sum_{m, n, c} \mathbb{F}^{c}(m, n)*\mathbb{F}^{c}(m-p, n-q)  + \sum_{m, n, c} \|\mathbb{F}^{c}(m-p, n-q)\|^2}{\sum_{m, n, c} \|\mathbb{F}^{c}(m, n)\|^2}
\label{eq:impldetail}    
\end{equation}

% \begin{equation}
% \begin{split}
% {
%     \mathbf{s}(p,q) & = -\frac{\sum_{m, n, c} \|\mathbb{F}^{c}(m, n) - \mathbb{F}^{c}(m-p, n-q)\|^2}{\sum_{m, n, c} \|\mathbb{F}^{c}(m, n)\|^2} \\
%     & = -\frac{\sum_{m, n, c} \|\mathbb{F}^{c}(m, n)\|^2 - 2*\sum_{m, n, c} \mathbb{F}^{c}(m, n)*\mathbb{F}^{c}(m-p, n-q)  + \sum_{m, n, c} \|\mathbb{F}^{c}(m-p, n-q)\|^2}{\sum_{m, n, c} \|\mathbb{F}^{c}(m, n)\|^2}
% }
% \end{split}
% \label{eq:selfsim_per_shift}    
% \end{equation}
% \begin{equation}
% \label{eq:style_loss}
% \begin{split}
% {
%  \mathcal{L}_{style_{out}} = &\sum_{p=0}^{P-1}{\frac{1}{C_{p} C_{p}} {\Big|\Big|K_p\big({{\big(\Psi}_{p}^{\mathbf{I}_{out}}\big)}^{\intercal}{{\big(\Psi}_{p}^{\textbf{I}_{out}}\big)} - \\ &{\big({\Psi}_{p}^{\mathbf{I}_{target}}\big)}^{\intercal}{\big({\Psi}_{p}^{\mathbf{I}_{target}}\big)\big)}\Big|\Big|}_{1} }
%  }
%  \end{split}
%  \end{equation}

Here, $m\in[max(0, p), min(p+H,H)]$ and $n\in[max(0,q), min(q+W,W)]$ indicate the overlapping region between current $(p, q)$-shifted copy and the original copy. $\|\mathbb{F}\|_2$ is the L2 norm of $\mathbb{F}$. The dominator $\sum_{m, n, c} \|\mathbb{F}_{m, n}^{c}\|^2$ is used for denormalization such that the scale of $\mathbf{s}(p,q)$ is independent of the scale of $\mathbb{F}$.

\textbf{Implementation Details}. $\sum_{m, n, c} \|\mathbb{F}^{c}(m, n)\|^2$ can be computed by using $\mathbb{F}^2$ as convolution input and a convolution filter with weights being all 1s and biases being all 0s. $\sum_{m, n, c} \mathbb{F}^{c}(m, n)*\mathbb{F}^{c}(m-p, n-q)$ can be computed by using the zero-padded $\mathbb{F}$, with $H/2$ zero padding on top and bottom sides and $W/2$ zero padding on left and right sides as convolution input and $\mathbb{F}$ as convolution filter. Similarly, $\sum_{m, n, c} \|\mathbb{F}^{c}(m-p, n-q)\|^2$ can be computed by using a $(2H, 2W)$ map, with the center region $[H/2:H/2+H, W/2:W/2+W)$ being 1 and other region being 0, as convolution input and $\mathbb{F}$ as convolution filter.

\subsection{Transposed Convolution Block}

Table~\ref{tab:transconvdiff} lists the main differences between typical transposed convolution operation and our transposed convolution operation.

\begin{figure}
    \centering
    \includegraphics[width=1.0\textwidth]{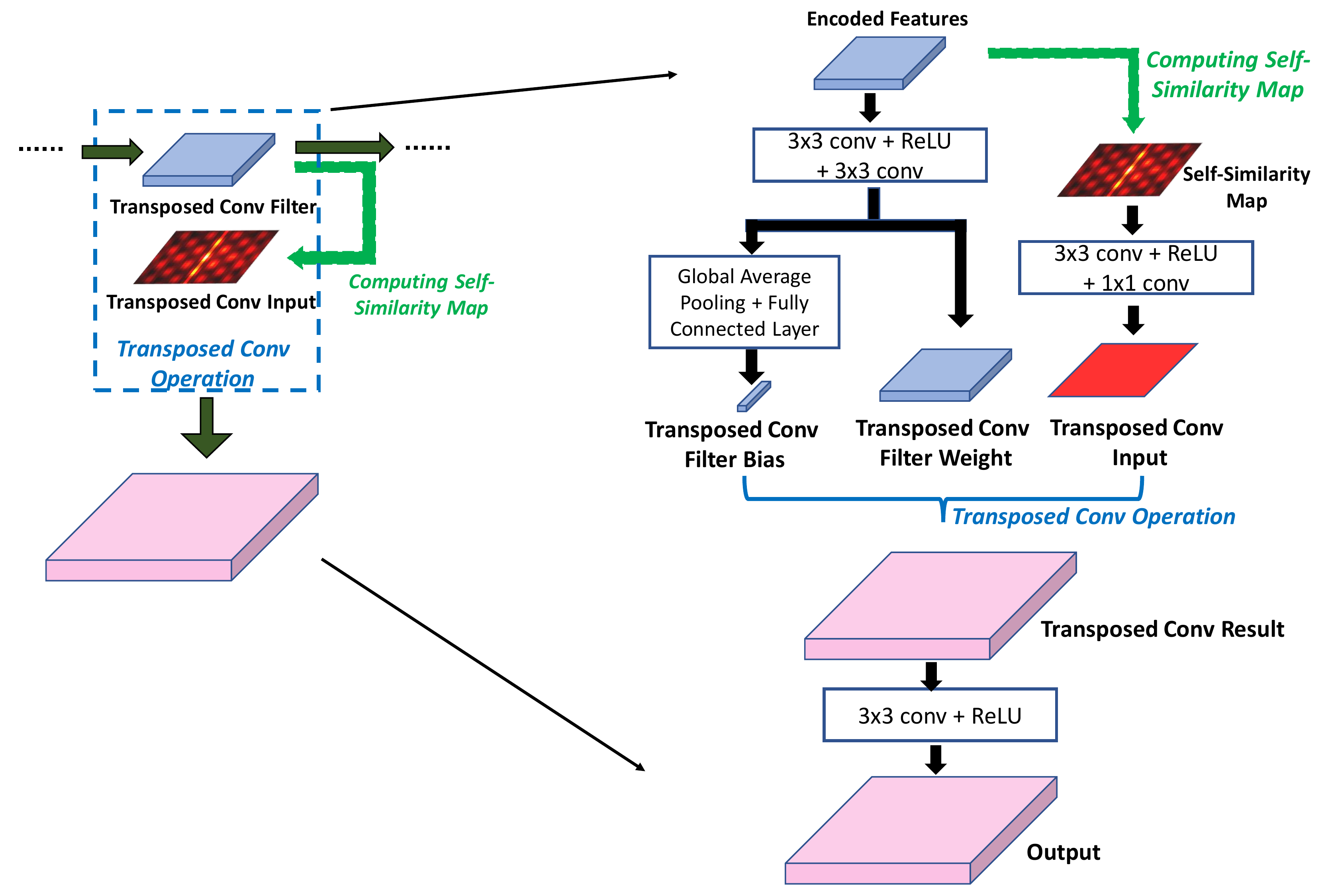}
    \caption{The details of a transposed convolution block. The left part shows the corresponding preview which is used in the Figure 3 in the main paper; the right part shows the details of this transposed convolution block.}
    \label{fig:transconv_block}
\end{figure}

\noindent Fig.~\ref{fig:transconv_block} shows the details for transposed convolution block in our framework. 

\begin{table}[h]
    \centering
    \scalebox{1.0}{
    \begin{tabular}{l|l|l}
    \multicolumn{1}{c}{} & typical transposed conv operation & our transposed conv operation \\
    \hline
    input & output from previous layer & self-similarity map from encoded features \\
    filter & learn-able parameters & feature maps from encoder \\
    bias term & learn-able parameters & avg-pooling of encoded features with linear transform \\
    filter size & small (e.g. 4x4, 3x3) & large (e.g. 8x8, 16x16, 32x32, 64x64) \\
    stride & 2(for upsampling purpose) & 1 \\
    \end{tabular}
    }
    \addtolength{\tabcolsep}{3pt}
    \vspace{2mm}
    \caption{Main differences between typical transposed convolution and our transposed convolution operation}
    \label{tab:transconvdiff}
    \vspace{-2mm}
\end{table}

\subsection{Network Details}
Table~\ref{tab:net_detail} shows the details of generator. The discriminator network is the same with pix2pixHD~\cite{wang2018high}. We use partial convolution based padding~\cite{liu2018partial} instead of zero padding for all the convolution layers.

\begin{table}
\centering
\scalebox{1.0}{
    \begin{tabular}{l|ccccc}
    \multicolumn{1}{c}{} & \multicolumn{1}{|c}{SSIM} & FID & c-FID & LPIPS & c-LPIPS \\
    \hline
    \hline
    % Naive tiling &  &  &  &  &  \\
    Self-tuning & 0.3157 & 95.829 & 0.4393 & 0.4078 & 0.3653 \\
    Non-station. & 0.3349 & 120.245 & 1.6888 & 0.4226 & 0.3911 \\
    sinGAN & 0.3270 & 147.9333  & 1.3806 & 0.4230 &  0.3829 \\
    pix2pixHD  & 0.3253 & 131.655 &  0.5472 & 0.4193 & 0.3780 \\
    \textbf{Ours} & 0.4533 & \textbf{78.4808} & \textbf{0.3973}  & \textbf{0.3246} & \textbf{0.3563} \\
    \hline
    Non-station.$^{*}$ & \textbf{0.4915} & 211.0645 & 1.4274 & 0.3411 & 0.3893 \\
    sinGAN$^{*}$ & 0.2913 & 154.651  & 1.6909 & 0.4787 & 0.4364 \\
    DeepTexture & 0.3011 & 82.053 & 0.5649 & 0.4175 & 0.3830 \\
    WCT  & 0.3124 & 144.208 & 0.4125 & 0.4427 & 0.4068 \\
    % Ground truth &  &  &  &  & \\
    \end{tabular}
}
% \vspace{2mm}
\caption{256 to 512 synthesis scores for different approaches averaged over 200 images. Non-station.$^{*}$, sinGAN$^{*}$, DeepTexture and WCT directly take the ground truth images as inputs.}
\label{tab:eval256to512}
\end{table}

\begin{table}
    \centering
    \vspace{-0.1cm}
    \scalebox{0.58}{
    \begin{tabular}{c|c|c|c|c|c|c}
         & Block & Filter Size & \# Filters & Stride/Up Factor & Sync BN & ReLU\\
        \hline
        \hline
         \multirow{9}{*}{Encoder} & Conv1& 3$\times$3& 3 $\rightarrow$ 64& 1& Y& Y\\
         & Conv2\_1& 3$\times$3& 64 $\rightarrow$ 128& 2& Y& Y\\
         & Conv2\_2& 3$\times$3& 128 $\rightarrow$ 128& 1& Y& Y\\
         & Conv3\_1& 3$\times$3& 128 $\rightarrow$ 256& 2& Y& Y\\
         & Conv3\_2& 3$\times$3& 256 $\rightarrow$ 256& 1& Y& Y\\
         & Conv4\_1& 3$\times$3& 256 $\rightarrow$ 512& 2& Y& Y\\
         & Conv4\_2& 3$\times$3& 512 $\rightarrow$ 512& 1& Y& Y\\
         & Conv5\_1& 3$\times$3& 512 $\rightarrow$ 1024& 2& Y& Y\\
         & Conv5\_2& 3$\times$3& 1024 $\rightarrow$ 1024& 1& Y& Y\\
         \hline
         \hline
        % \multirow{7}{*}{TransConv\_Block3 \newline (w/ Conv3\_2)}  
        & FilterBranch\_Conv1 & 3$\times$3 & 256 $\rightarrow$ 256 & 1 & - & Y \\
        & FilterBranch\_Conv2 & 3$\times$3 & 256 $\rightarrow$ 256 & 1 & - & - \\
        & FilterBranch\_FC1 & - & 256 $\rightarrow$ 256 & 1 & - & - \\
        TransConv\_Block3 & SelfSimilarityMapBranch\_Conv1 & 3$\times$3 & 1 $\rightarrow$ 8 & 1 & - & Y \\
        (w/ Conv3\_2) & SelfSimilarityMapBranch\_Conv2 & 3$\times$3 & 8 $\rightarrow$ 1 & 1 & - & - \\
        & transposed Convolution Operation & $\frac{\text{orig\_H}}{4}\times \frac{\text{orig\_W}}{4}$ & filter: 256, input: 1 $\rightarrow$ 256 & - & - & - \\
        & OutputBranch\_Conv & 3$\times$3 & 256 $\rightarrow$ 256 & 1 & - & Y \\
        \hline
        % \multirow{7}{*}{TransConv\_Block4 \newline (w/ Conv4\_2)}  
        & FilterBranch\_Conv1 & 3$\times$3 & 512 $\rightarrow$ 512 & 1 & - & Y \\
        & FilterBranch\_Conv2 & 3$\times$3 & 512 $\rightarrow$ 512 & 1 & - & - \\
        TransConv\_Block4 & FilterBranch\_FC1 & - & 512 $\rightarrow$ 512 & 1 & - & - \\
        (w/ Conv4\_2) & SelfSimilarityMapBranch\_Conv1 & 3$\times$3 & 1 $\rightarrow$ 8 & 1 & - & Y \\
        & SelfSimilarityMapBranch\_Conv2 & 3$\times$3 & 8 $\rightarrow$ 1 & 1 & - & - \\
        & transposed Convolution Operation & $\frac{\text{orig\_H}}{8}\times \frac{\text{orig\_W}}{8}$ & filter: 512, input: 1 $\rightarrow$ 512 & - & - & - \\
        & OutputBranch\_Conv & 3$\times$3 & 512 $\rightarrow$ 512 & 1 & - & Y \\
        \hline
        % \multirow{5}{*}{TransConv\_Block5 (w/ Conv5\_2)}  
        & FilterBranch\_Conv1 & 3$\times$3 & 1024 $\rightarrow$ 1024 & 1 & - & Y \\
        & FilterBranch\_Conv2 & 3$\times$3 & 1024 $\rightarrow$ 1024 & 1 & - & - \\
        TransConv\_Block5 & FilterBranch\_FC1 & - & 1024 $\rightarrow$ 1024 & 1 & - & - \\
        (w/ Conv5\_2) & SelfSimilarityMapBranch\_Conv1 & 3$\times$3 & 1 $\rightarrow$ 8 & 1 & - & Y \\
        & SelfSimilarityMapBranch\_Conv2 & 3$\times$3 & 8 $\rightarrow$ 1 & 1 & - & - \\
        & transposed Convolution Operation & $\frac{\text{orig\_H}}{16}\times \frac{\text{orig\_W}}{16}$ & filter: 1024, input: 1 $\rightarrow$ 1024 & - & - & - \\
        & OutputBranch\_Conv & 3$\times$3 & 1024 $\rightarrow$ 1024 & 1 & - & Y \\
        \hline
        \hline
         \multirow{11}{*}{Decoder} & BilinearUpSample1 \newline (w/ TransConv\_Block5 output)& - & - & 2 & - & -\\
         & Conv6& 3$\times$3& 1024 $\rightarrow$ 512& 1& Y& Y\\
         & Sum (Conv6 + TransConv\_Block4 output) & - & - &  - & - & -\\
         & BilinearUpSample2 & - & - & 2 & - & -\\
         & Conv7& 3$\times$3& 512 $\rightarrow$ 256& 1& Y& Y\\
         & Sum (Conv7 + TransConv\_Block3 output) & - & - &  - & - & -\\
         & BilinearUpSample3 & - & - & 2 & - & -\\
         & Conv8& 3$\times$3& 256 $\rightarrow$ 128& 1& Y& Y\\
         & BilinearUpSample4 & - & - & 2 & - & -\\
         & Conv9& 3$\times$3& 128 $\rightarrow$ 64& 1& Y& Y\\
         & Conv10& 3$\times$3& 64 $\rightarrow$ 3& 1 & - & -\\
         \hline
    \end{tabular}
    }
    \caption{The details of network parameters. TransConv\_Block3-5 represent the three transposed convolution blocks in our framework (The diagrams can be found in Figure 2 in the main paper). SyncBatchNorm column indicates Synchronized Batch Normalization layer after Conv. ReLU column shows whether ReLU is used (following the SyncBatchNorm if SyncBatchNorm is used). BilinearUpSample represents bilinear upsampling. Sum denotes the simple summation. \text{orig\_H} and \text{orig\_W} are input image's height and width.}
    \label{tab:net_detail}
    \vspace{-0.2cm}
\end{table}

% \bibliographystyle{ACM-Reference-Format}
% \bibliography{egbib}

\section{Additional Comparison}

\subsection{256 to 512 Synthesis}

In Table~\ref{tab:eval256to512}, we provide the quantitative comparisons for the synthesis results of 256 to 512. 

\subsection{128 to 256 Synthesis}

\noindent\textbf{Non-stat. and Non-stat.$^{*}$ baselines}: we take the original code from the author's github repository. The original training strategy for each training iteration is: 1). randomly crop a $2H \times 2W$ from the original big image($>2H \times 2W$) as the target image; 2). from the target image, randomly crop a $H \times W$ image as the input image. Thus, for 128 to 256 synthesis, to train Non-stat. (without seeing ground truth $256 \times 256$ image), for each training iteration, we randomly crop a $96 \times 96$ image from the input $128 \times 128$ image as target image then from the target image, we randomly crop a $48 \times 48$ image as input.  To train Non-stat.* (with directly seeing ground truth $256 \times 256$ image), for each training  iteration, we randomly crop a $128 \times 128$ image from the ground truth $256 \times 256$ image as the target image and then from the target image, we randomly crop a $64 \times 64$ image as input image. For both Non-stat. and Non-stat.*, the inference stage will take $128 \times 128$ image as input.

\noindent\textbf{sinGAN and sinGAN$^{*}$ baselines}: for training with sinGAN, we used the original author's implementation available on github. And we used the default settings the author provided in their source code. sinGAN code can synthesize textures in two different modes, one that generates a random variation which is of the same size as input texture (we directly using ground truth $256\times256$ for training, denoted as sinGAN$^{*}$), and another that generates a texture of larger size (only using $128\times128$ image, denoted as sinGAN).

\begin{figure}
\centering
\scalebox{0.7}{
% \addtolength{\tabcolsep}{-4pt}   
\begin{tabular}{c|cccccccc}
\multicolumn{1}{c}{} & \multicolumn{8}{c}{} \\
input & ours & Self-tuning & pix2pixHD & WCT & sinGAN$^{*}$ & Non-stat.$^{*}$ & DeepTexture$^{*}$ & ground truth \\
\hline
\hline
\includegraphics[width=0.07\textwidth]{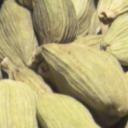} &
\includegraphics[width=0.14\textwidth]{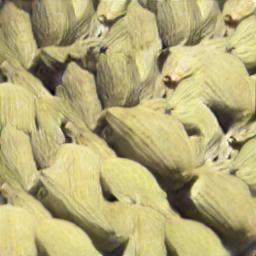} &
\includegraphics[width=0.14\textwidth]{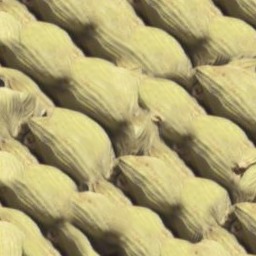} &
\includegraphics[width=0.14\textwidth]{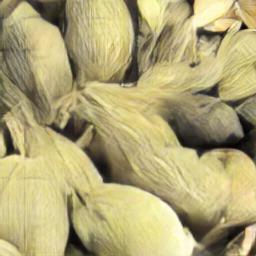} &
\includegraphics[width=0.14\textwidth]{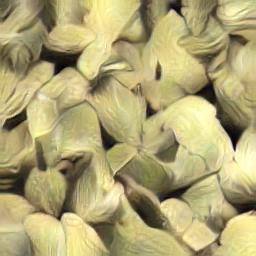} &
\includegraphics[width=0.14\textwidth]{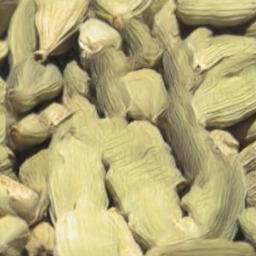} &
\includegraphics[width=0.14\textwidth]{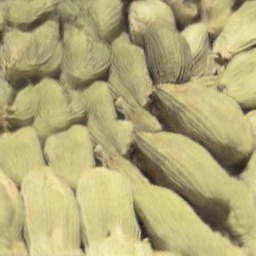} &
\includegraphics[width=0.14\textwidth]{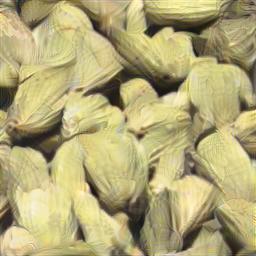} &
\includegraphics[width=0.14\textwidth]{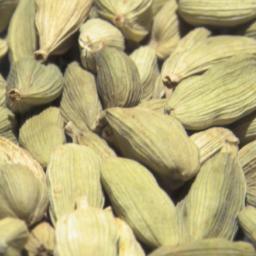} \\
\includegraphics[width=0.07\textwidth]{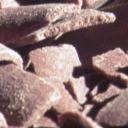} &
\includegraphics[width=0.14\textwidth]{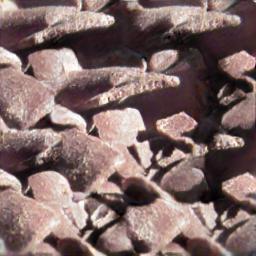} &
\includegraphics[width=0.14\textwidth]{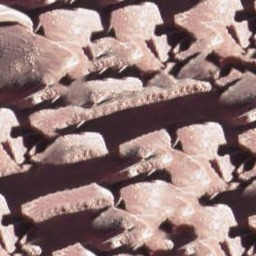} &
\includegraphics[width=0.14\textwidth]{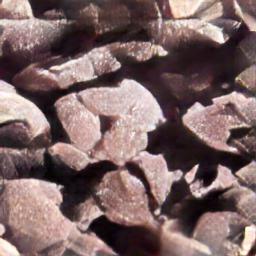} &
\includegraphics[width=0.14\textwidth]{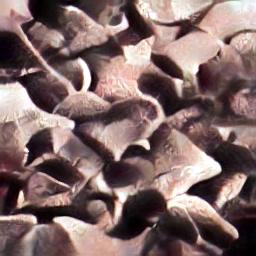} &
\includegraphics[width=0.14\textwidth]{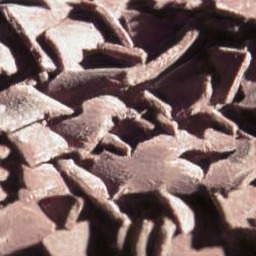} &
\includegraphics[width=0.14\textwidth]{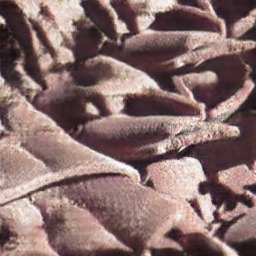} &
\includegraphics[width=0.14\textwidth]{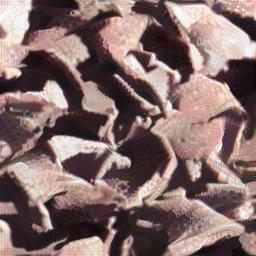} &
\includegraphics[width=0.14\textwidth]{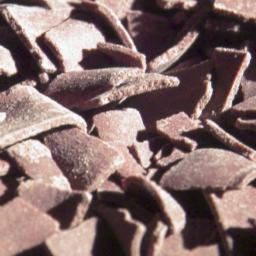} \\
\includegraphics[width=0.07\textwidth]{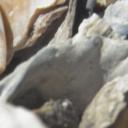} &
\includegraphics[width=0.14\textwidth]{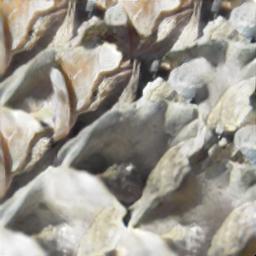} &
\includegraphics[width=0.14\textwidth]{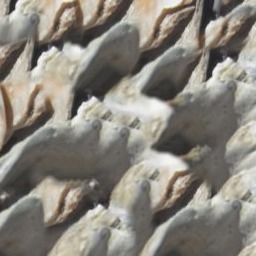} &
\includegraphics[width=0.14\textwidth]{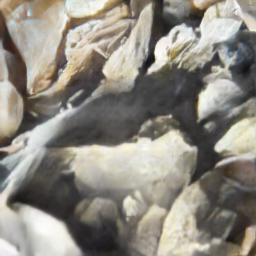} &
\includegraphics[width=0.14\textwidth]{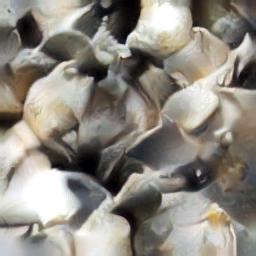} &
\includegraphics[width=0.14\textwidth]{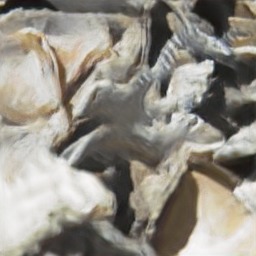} &
\includegraphics[width=0.14\textwidth]{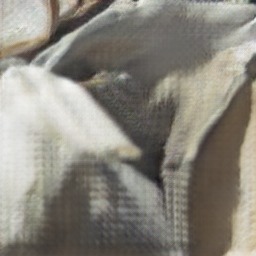} &
\includegraphics[width=0.14\textwidth]{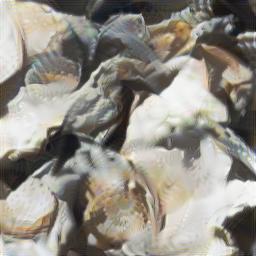} &
\includegraphics[width=0.14\textwidth]{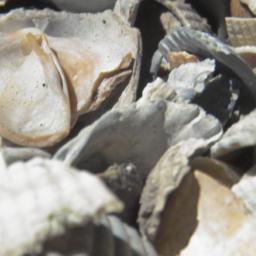} \\
\includegraphics[width=0.07\textwidth]{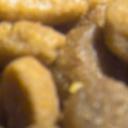} &
\includegraphics[width=0.14\textwidth]{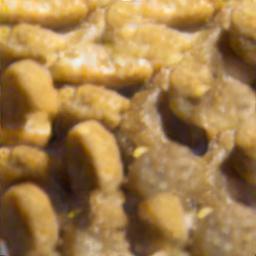} &
\includegraphics[width=0.14\textwidth]{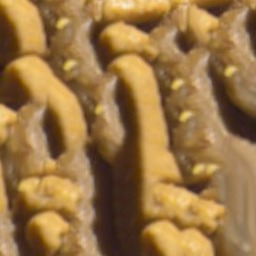} &
\includegraphics[width=0.14\textwidth]{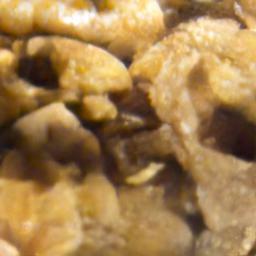} &
\includegraphics[width=0.14\textwidth]{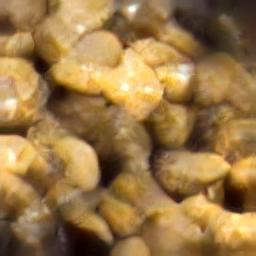} &
\includegraphics[width=0.14\textwidth]{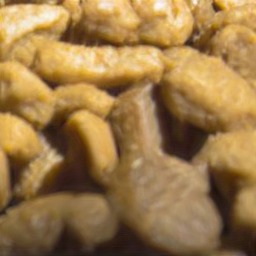} &
\includegraphics[width=0.14\textwidth]{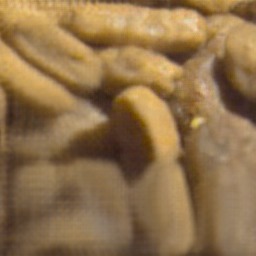} &
\includegraphics[width=0.14\textwidth]{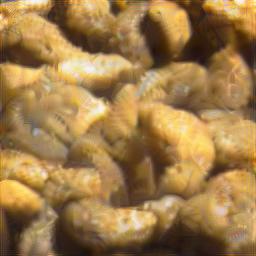} &
\includegraphics[width=0.14\textwidth]{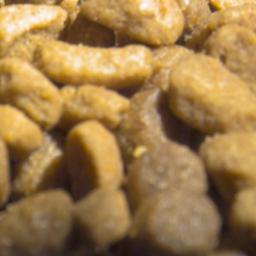} \\
\end{tabular}
}
\caption{Results of different approaches on 128 to 256 texture synthesis. sinGAN$^{*}$ Non-stat.$^{*}$ show the results of training with directly seeing the ground truth at target size. (Training with ground truth means using the ground truth 256$\times$256 image as the target for each training iteration.) WCT is the style transfer based method. DeepTexture directly takes ground truth images as inputs.}
\label{fig:result256_suppl}
\end{figure}

% comparison images
\begin{figure}
\centering
\scalebox{0.75}{
\addtolength{\tabcolsep}{-4pt}   
\begin{tabular}{c|cccccccc}
\multicolumn{1}{c}{} & \multicolumn{8}{c}{} \\
input & ours & Self-tuning & pix2pixHD & WCT & SinGAN & Non-stat. & DeepTexture$^{*}$ & ground truth \\
\hline
\hline

\includegraphics[width=0.07\textwidth]{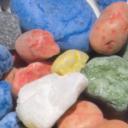} &
\includegraphics[width=0.14\textwidth]{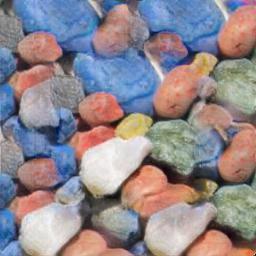} &
\includegraphics[width=0.14\textwidth]{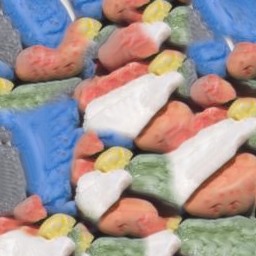} &
\includegraphics[width=0.14\textwidth]{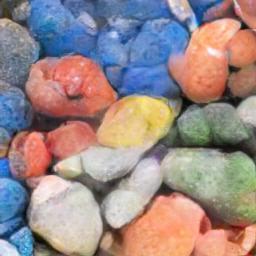} &
\includegraphics[width=0.14\textwidth]{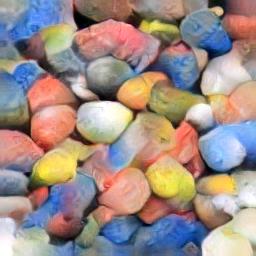} &
\includegraphics[width=0.14\textwidth]{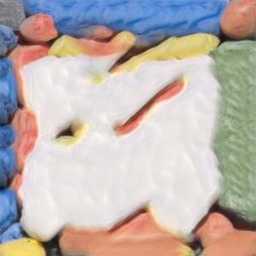} &
\includegraphics[width=0.14\textwidth]{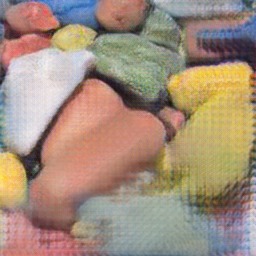} &
\includegraphics[width=0.14\textwidth]{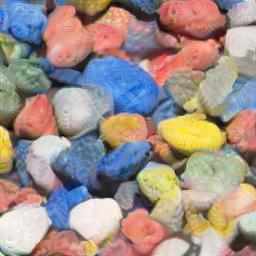} &
\includegraphics[width=0.14\textwidth]{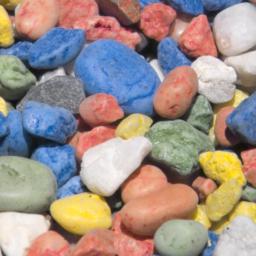} \\
\includegraphics[width=0.07\textwidth]{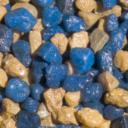} &
\includegraphics[width=0.14\textwidth]{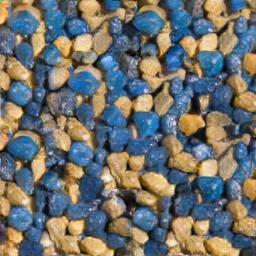} &
\includegraphics[width=0.14\textwidth]{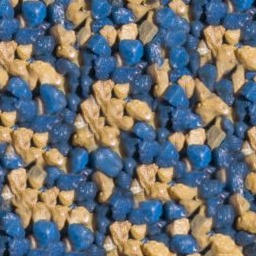} &
\includegraphics[width=0.14\textwidth]{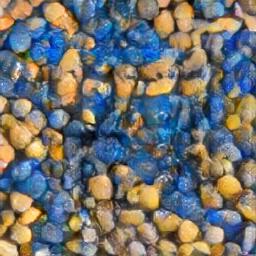} &
\includegraphics[width=0.14\textwidth]{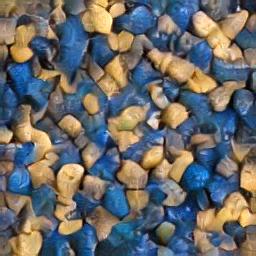} &
\includegraphics[width=0.14\textwidth]{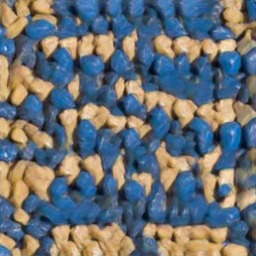} &
\includegraphics[width=0.14\textwidth]{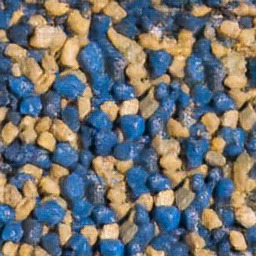} &
\includegraphics[width=0.14\textwidth]{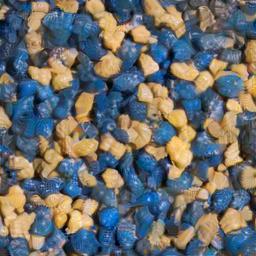} &
\includegraphics[width=0.14\textwidth]{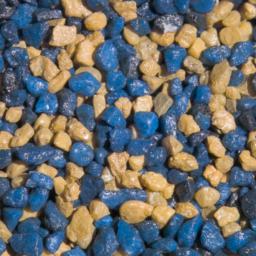} \\
\includegraphics[width=0.07\textwidth]{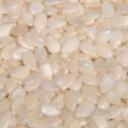} &
\includegraphics[width=0.14\textwidth]{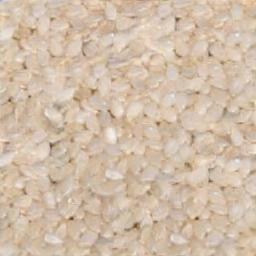} &
\includegraphics[width=0.14\textwidth]{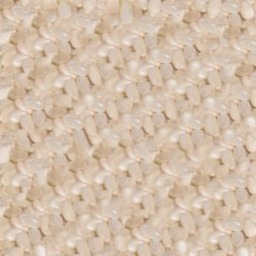} &
\includegraphics[width=0.14\textwidth]{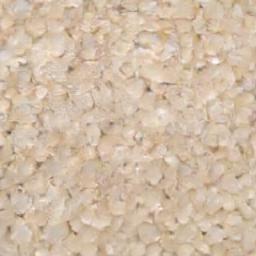} &
\includegraphics[width=0.14\textwidth]{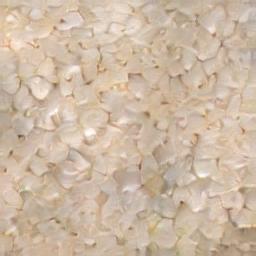} &
\includegraphics[width=0.14\textwidth]{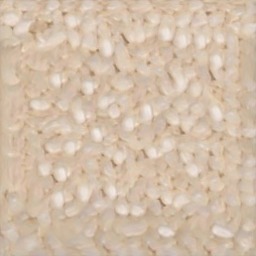} &
\includegraphics[width=0.14\textwidth]{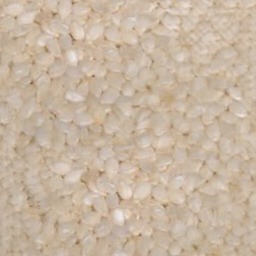} &
\includegraphics[width=0.14\textwidth]{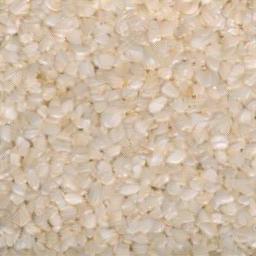} &
\includegraphics[width=0.14\textwidth]{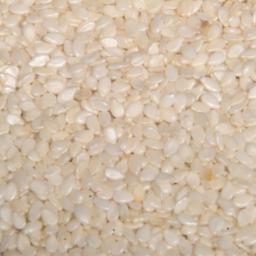} \\
\includegraphics[width=0.07\textwidth]{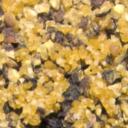} &
\includegraphics[width=0.14\textwidth]{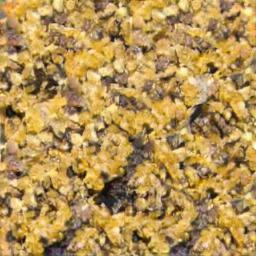} &
\includegraphics[width=0.14\textwidth]{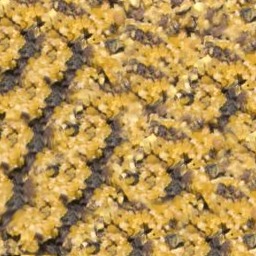} &
\includegraphics[width=0.14\textwidth]{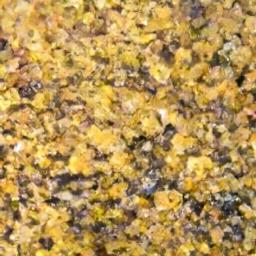} &
\includegraphics[width=0.14\textwidth]{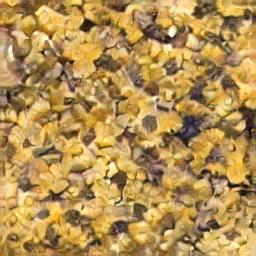} &
\includegraphics[width=0.14\textwidth]{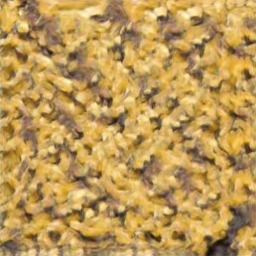} &
\includegraphics[width=0.14\textwidth]{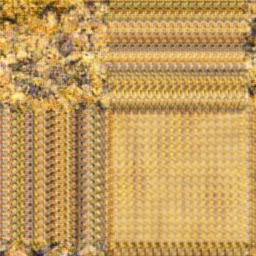} &
\includegraphics[width=0.14\textwidth]{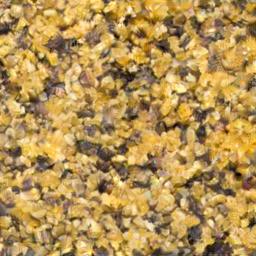} &
\includegraphics[width=0.14\textwidth]{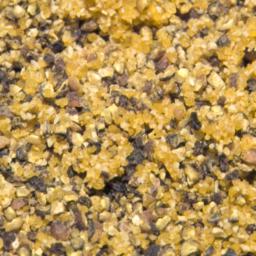} \\
\includegraphics[width=0.07\textwidth]{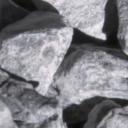} &
\includegraphics[width=0.14\textwidth]{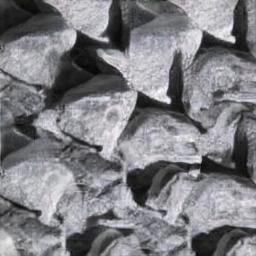} &
\includegraphics[width=0.14\textwidth]{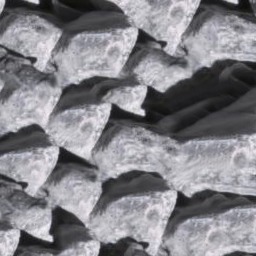} &
\includegraphics[width=0.14\textwidth]{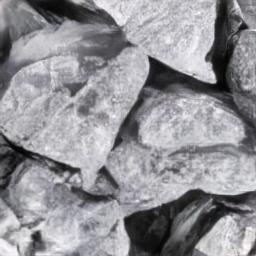} &
\includegraphics[width=0.14\textwidth]{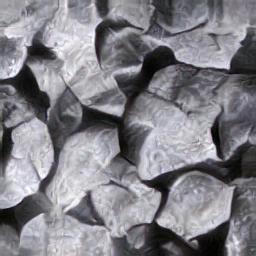} &
\includegraphics[width=0.14\textwidth]{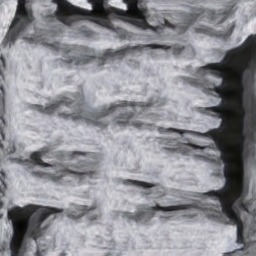} &
\includegraphics[width=0.14\textwidth]{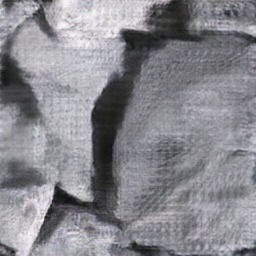} &
\includegraphics[width=0.14\textwidth]{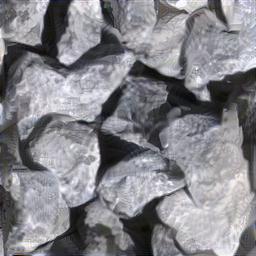} &
\includegraphics[width=0.14\textwidth]{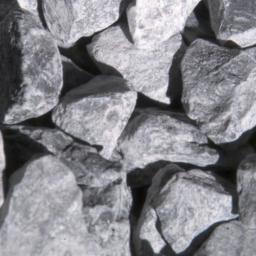} \\

\includegraphics[width=0.07\textwidth]{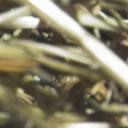} &
\includegraphics[width=0.14\textwidth]{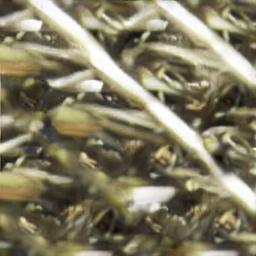} &
\includegraphics[width=0.14\textwidth]{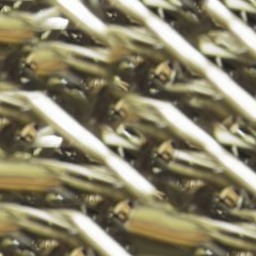} &
\includegraphics[width=0.14\textwidth]{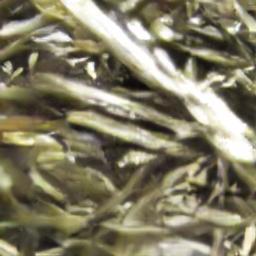} &
\includegraphics[width=0.14\textwidth]{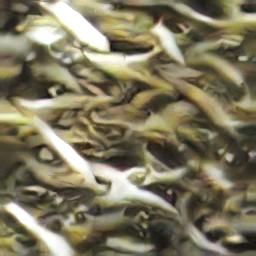} &
\includegraphics[width=0.14\textwidth]{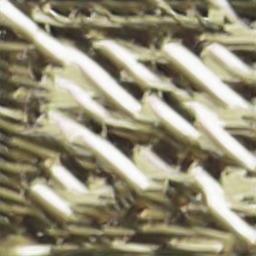} &
\includegraphics[width=0.14\textwidth]{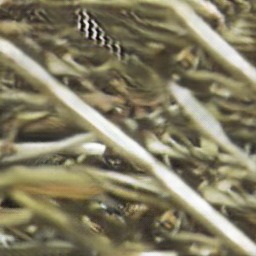} &
\includegraphics[width=0.14\textwidth]{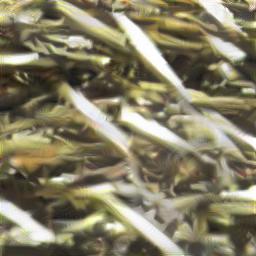} &
\includegraphics[width=0.14\textwidth]{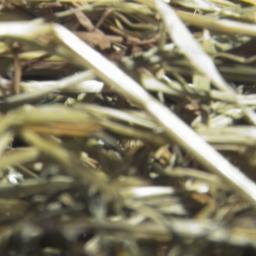} \\

\includegraphics[width=0.07\textwidth]{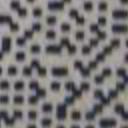} &
\includegraphics[width=0.14\textwidth]{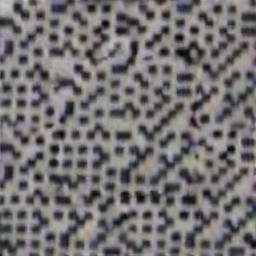} &
\includegraphics[width=0.14\textwidth]{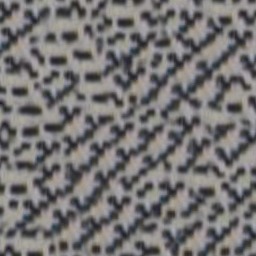} &
\includegraphics[width=0.14\textwidth]{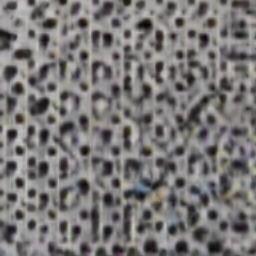} &
\includegraphics[width=0.14\textwidth]{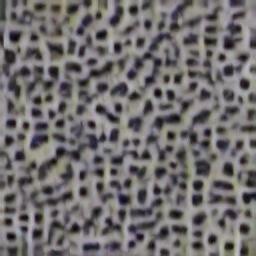} &
\includegraphics[width=0.14\textwidth]{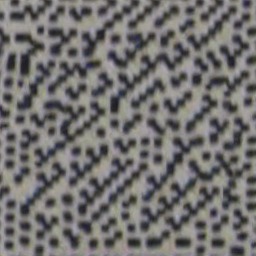} &
\includegraphics[width=0.14\textwidth]{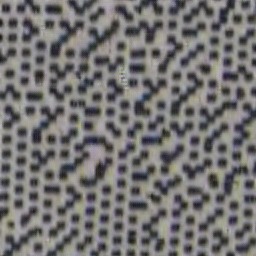} &
\includegraphics[width=0.14\textwidth]{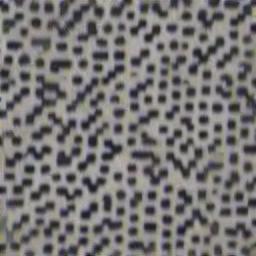} &
\includegraphics[width=0.14\textwidth]{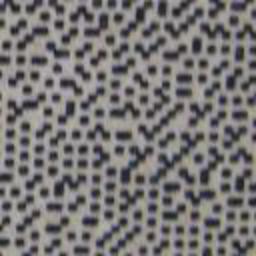} \\

\end{tabular}
}
\caption{Results of different approaches on 128 to 256 texture synthesis. SinGAN and Non-stat. results show the results of training without directly seeing ground truth at the exact target size.}
\label{fig:result256_suppl2}
\end{figure}

\end{document}

%% file: sec_intro.tex
%%%%%%%%% BODY TEXT
\section{Introduction}
\label{sec:intro}

Texture synthesis is defined as the problem of generating a large image output given a small example input such that the visual features and structures are preserved both locally and globally. Many methods have been explored in the past two decades including pixel-based methods~\cite{efros1999texture}, assembling based methods~\cite{efros2001image,kwatra2003graphcut}, optimization based methods~\cite{kwatra2005texture,kaspar2015self}, etc.

Inspired by the unprecedented success of deep learning in computer vision, others have explored deep learning methods for texture synthesis. Existing works fall into one of two categories. Either an optimization procedure is used to match deep feature statistics in a pre-trained network~\cite{gatys2015texture,yijun2017universal}, resulting in a slow generation process; or a network is trained to overfit on a fixed image or set of images~\cite{li2017diversified,zhou2018non,shaham2019singan}, which prevents it from generalizing to unseen textures and needs to spend huge re-training time for every unseen texture image.

One reason for the bad generalization ability of these aforementioned methods~\cite{li2017diversified,zhou2018non,shaham2019singan} is because these one-model-per-image (set) approaches usually employ conventional image-to-image translation networks, which first embed the input into a feature space and then fully rely on a sequence of upsampling and convolutional layers to reach the target output size. 
% One reason for bad generalization by the aforementioned neural networks methods may occur because these one-model-per-image(set) approaches usually employ some conventional image(noise)-to-image translation networks, which first embed the input into feature space then rely on a sequence of upsampling layers and convolution layers to progressively upsample the features to the target output size. 
Each upsampling and convolutional layer is a local operation lacking of global awareness. This design works well for tasks such as image super resolution, where the task is to enhance or modify local details. However, texture synthesis differs from super resolution in that texture synthesis, when viewed from a classical perspective, involves displacing and assembling copies of the input texture using different optimal offsets in a seamless way. The optimal displacement and assembling strategy involves much longer-range operations and compatibility checking, which are usually not easy to model with the conventional design by fully relying on a sequence of local up/down-sampling and (de)convolutional layers. 

In the column pix2pixHD of Figure~\ref{fig:teaser_comp}, we show that a conventional image-to-image translation network adapted from pix2pixHD~\cite{wang2018high} fails to perform reasonable texture synthesis, but instead mostly just enlarges the local contents for the input textures even though it has been trained to convergence using the same input and output pairs as our method. 

In this paper, we propose a new deep learning based texture synthesis framework that generalizes to arbitrary unseen textures and synthesizes larger-size outputs. From a classical view, the texture synthesis task can also be interpreted as the problem of first finding an appropriate offset to place a copy of the input texture image and then using optimization technique to find the optimal seam between this newly placed copy and the existing image to assemble them together. 
% One classical way of interpreting texture synthesis task would be first finding an appropriate offset to place a copy of the input texture image and then using optimization technique to find the optimal seam between this newly placed copy and the existing image to assemble them together. 
% Our method is partly inspired by the traditional assembling (stitching)-based methods, which first find an appropriate offset to place a copy of the input texture image and then use optimization technique to find the optimal seam between this newly placed copy and the existing image to assemble them together. 
Our method follows some similar spirits but have some major differences
%We differ from this method 
in the following manner: 1) We perform assembling in feature space and at multiple scales; 2) The optimal shifting offset and assembling weights are modeled with the help of a score map, which captures the similarity and correlation between different regions of the encoded texture image. We call this score map a \textit{self-similarity map} (discussed in details in Section~\ref{sec:approach}); 3) We later show that the shifting and assembling operations can be efficiently performed with a single forward pass of a \textit{transposed convolution operation}~\cite{dumoulin2016guide}, where we \textit{directly use the encoded feature of input textures as transposed convolution filters, and the self-similarity map as transposed convolution input}. Unlike traditional transposed convolution, our transposed convolution filters are not learnable parameters. While self-similarity map plays a key role in preserving the regular structural patterns, alternately, our framework also allows to take random noise map as input instead of self-similarity map to generate diverse outputs and arbitrarily large texture output with a single shot by accordingly sampling large random noise map for irregular structural texture inputs.

In this work, we make the following contributions: 1) We present a generalizable texture synthesis framework that performs faithful synthesis on unseen texture images in nearly real time with a single forward pass. 2) We propose a self-similarity map that captures the similarity and correlation information between different regions of a given texture image. 3) We show that the shifting and assembling operations in traditional texture synthesis methods can be efficiently implemented using a transposed convolution operation. 4) We achieve state-of-the-art texture synthesis quality as measured by existing image metrics, metrics designed specifically for texture synthesis, and in user study. 5) We show that our framework is also able to generate diverse and arbitrarily large texture synthesis results by sampling random noise maps.

%% file: sec_relatedwork.tex
\section{Related Work}

We provide a brief overview of the existing texture synthesis methods. A complete texture synthesis survey can be found in~\cite{wei2009state}, which is out of the scope of this work.

\textbf{Non-parametric Texture Synthesis}. Existing texture synthesis methods include pixel-based methods~\cite{efros1999texture,wei2000fast}, assembling based methods~\cite{efros2001image,liang2001real,kwatra2003graphcut,pritch2009shift}, optimization based methods~\cite{portilla2000parametric,kwatra2005texture,rosenberger2009layered,kaspar2015self}, appearance space synthesis~\cite{lefebvre2006appearance}, etc. There are also some other works~\cite{hertzmann2001image,zhang2003synthesis,wu2004feature,lefebvre2006appearance,rosenberger2009layered,wu2013optimized} showing interesting synthesis results; however, they usually need additional user manual inputs. 

Among these traditional methods, self-tuning texture optimization~\cite{kaspar2015self} is the current state-of-the-art method. It uses image melding~\cite{darabi2012image} with automatically generated and weighted guidance channels, which helps to reconstruct the middle-scale structures in the input texture. Our method is motivated by assembling based methods. \cite{kwatra2003graphcut} is a representative method of this kind, where texture synthesis is formulated as a graph cut problem. The optimal offset for displacing the input patch and the optimal cut between the patches can be found by solving the graph cut objective function, which sometimes could be slow.

\textbf{Deep Feature Matching-based Texture Synthesis}. Traditional optimization based methods \cite{portilla2000parametric,kwatra2005texture,rosenberger2009layered,kaspar2015self} rely on matching the global statistics of the hand-crafted features defined on the input and output textures. 
% synthesize output by matching global image statistics such as histograms and wavelet coefficients between the input and output textures. However, the quality of generation largely depends on the hand-crafted features and objective functions.
%, and even the best classical method \cite{kaspar2015self} struggles to accurately synthesize complex textures.
Recently, some deep neural networks based methods have been proposed as a way to use the features learned from natural image priors to guide the optimization procedure. Gatys et al.~\cite{gatys2015texture} define the optimization procedure as minimizing the difference in gram matrices of the deep features between the input and output texture images. Sendik et al.~\cite{sendik2017deep} and Liu et al.~\cite{gang2016texture} modify the loss proposed in \cite{gatys2015texture} by adding a structural energy term and a spectrum constraint, respectively, to generate structured and regular textures. However, in all cases, these optimization-based methods are prohibitively slow due to the iterative optimizations.

\textbf{Learning-based Texture Synthesis}. Johnson et al.~\cite{johnson2016perceptual} and Ulyanov \cite{ulyanov2016texture} alleviate the previously mentioned optimization problem by training a neural network to directly generate the output, using the same loss as in \cite{gatys2015texture}. This setup moves the computational burden to training time, resulting in faster inference time. However, the learned network can only synthesize the texture it was trained on and cannot generalize to new textures.

A  more recent line of work \cite{zhou2018non,shaham2019singan,chuan2016precomputed,li2017diversified,fruhstuck2019tilegan,jetchev2016texture,bergmann2017learning,alanov2019user} has proposed using Generative Adversarial Networks (GANs) for more realistic texture synthesis while still suffering from the inability to generalize to new unseen textures.

Zhou et al.~\cite{zhou2018non} learn a generator network that expands $k \times k$ texture blocks into $2k \times 2k$ output through a combination of adversarial, $L_1$, and style (gram matrix) loss. Li et al.\ and Shaham et  al.~\cite{chuan2016precomputed,shaham2019singan} use a special discriminator that examines statistics of local patches in feature space. However, even these approaches can only synthesize a single texture which it has been trained on.

Other efforts \cite{li2017diversified,jetchev2016texture,bergmann2017learning,alanov2019user,fruhstuck2019tilegan} try to train on a set of texture images. During test time, the texture being generated is either chosen by the network \cite{jetchev2016texture,bergmann2017learning} or user-controlled \cite{li2017diversified,alanov2019user}. \cite{fruhstuck2019tilegan} propose a non-parametric method to synthesize large-scale, varied outputs by combining intermediate feature maps. However, these approaches limit generation to textures available in the training set, and thus are unable to produce unseen textures out of the training set.

Li et al.~\cite{yijun2017universal} apply a novel whitening and colorizing transform to an encoder-decoder architecture, allowing them to generalize to unseen textures, but rely on inner SVD decomposition which is slow. %However, the inner SVD decomposition is expensive to compute, especially for large textures, and thus cannot run in real-time; 
Additionally, it can only output texture images with the same size as the input.

Yu et al.~\cite{yu2019texture} perform the interpolation between two or more source textures. While forcing two source textures to be identical can convert it to the texture synthesis setting, it will
reduce the framework to be more like a conventional CNN. Besides probably suffering from the issues of conventional CNNs, its main operations of per-feature-entry shuffling, retiling and blending would greatly break the regular or large structure patterns in the input.

\textbf{Other Image-to-Image Tasks}. GANs~\cite{goodfellow2014generative} have also been used in other image-to-image tasks~\cite{isola2016image,wang2018high,dundar2020panoptic}. Ledig et al. \cite{ledig2016photo} used it to tackle the problem of super-resolution, where detail is added to a low-resolution image to produce high-definition output.  In contrast to these tasks, the texture synthesis problem is to synthesize new, varied regions similar to the input, and not to provide more details to an existing layout as in \cite{ledig2016photo} or translate the texture to a related domain as in \cite{isola2016image}. Other recipes like converting texture synthesis to an image inpainting problem usually cannot get satisfying results as they usually cannot handle big holes where we need to do the synthesis. 

% None of the prior work mentioned above can satisfy all the three desirable properties listed in Table~\ref{tab:intro-comparison} at the same time. 

\textbf{Similarity Map}. Our framework relies on computing the self-similarity map, which is similar in spirit to the deep correlation formulation in \cite{sendik2017deep}. The difference is that \cite{sendik2017deep} computes a dot product map between the feature map and its spatially shifted version and uses it as a regularizing term in their optimization objective; in contrast, we aggregate all the channels' information to compute a single-channel difference similarity map and use it to model the optimal synthesis scheme in the network with a single pass.

%% file: sec_approach.tex
\section{Our Approach}
\label{sec:approach}

\begin{figure*}[h]
    \centering
    \includegraphics[width=0.98\textwidth]{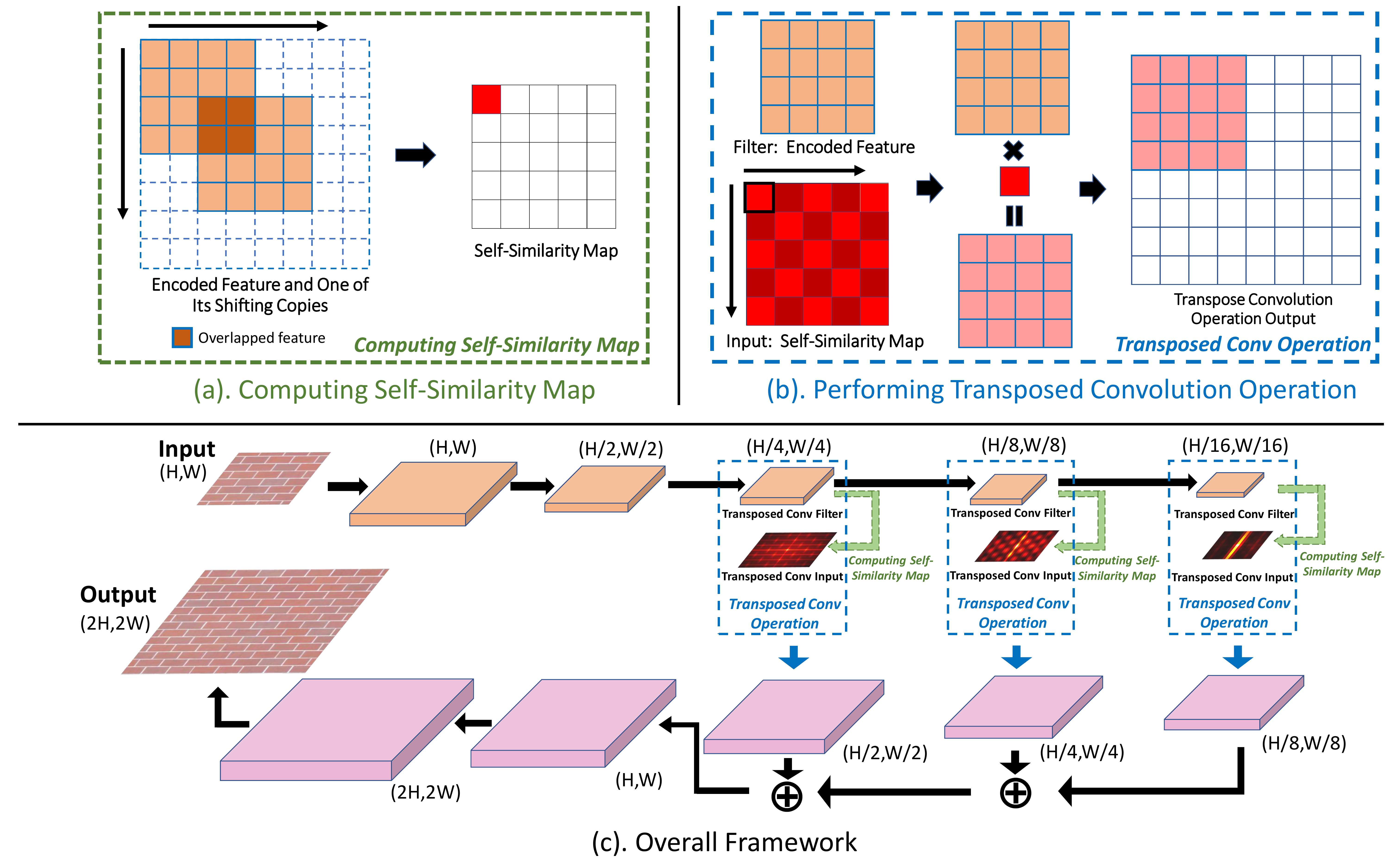}
    \caption{(a) shows how the self-similarity map is computed. (b) shows how to perform the transposed convolution operation. Both (a) and (b) are the components used in our overall framework (c), shown with \textcolor{Green}{green} and \textcolor{blue}{blue} colors, respectively. Full animation of (a) and (b) can be found in the supplementary video. In (c), yellow boxes represent the features in the encoder. The encoded features in the last three scales are first used to compute the self-similarity maps, as shown in (a). We then perform the transposed convolution operation as shown in (b), where encoded features are used as transposed convolution filters to convolve the self-similarity maps. The convolved outputs are then used in the decoder to generate the final image.}
    \label{fig:framework}
\end{figure*}

\textbf{Problem Definition}: Given an input texture patch, we want to expand the input texture to a larger output whose local pattern resembles the input texture pattern. Our approach to this problem shares some similar spirits with the traditional assembling based methods which try to find the optimal displacements of copies of the input texture, as well as the corresponding assembly scheme to produce a large, realistic texture image output. We will first formulate the texture expansion problem as a weighted linear combination of displaced deep features at various shifting positions, and then discuss how to use the transposed convolution operation to address it.

\textbf{Deep Feature Expansion}: Let $\mathbb{F} \in \mathbf{R}^{C\times H\times W}$ be the deep features of an input texture patch, with $C$, $H$ and $W$ being the number of channels, the height, and width, respectively. We create a spatially expanded feature map, for instance by a factor of 2, by simply pasting and accumulating $\mathbb{F}$ into a ${C\times 2H\times 2W}$ space. This is done by shifting $\mathbb{F}$ along the width axis with a progressive step ranging from 0 to $W$, as well as along the height axis with a step ranging from 0 to $H$. All the shifted maps are then aggregated together to give us an expanded feature map $\mathbb{G} \in \mathbf{R}^{C\times 2H\times 2W}$.

To aggregate the shifted copies of $\mathbb{F}$, we compute a weighted sum of them. For instance, to calculate the feature $G(i,j) \in \mathbf{R}^{C}$, we aggregate all possible shifted copies of $\mathbb{F}(\cdot,\cdot) \in \mathbf{R}^{C}$ that fall in the spatial location $(i,j)$. While previous approaches rely on hand crafted or other heuristics to aggregate the overlapping features, in our approach, we propose to weight each shifted feature map with a similarity score that quantifies the semantic distance between the original $\mathbb{F}$ and its shifted copy. Finally, aggregation is done by simple summation of the weighted features. Mathematically, $\mathbb{G}$ can be given by

% \begin{equation}
% \label{eq:paste_accumulate}
%     \mathbb{G}^{c} = \sum_{i,j} \sum_{p,q}R(\mathbf{s}(p,q)) \mathbb{F}^{c}_{p,q}(i,j)
% \end{equation}

\begin{equation}
\mathbb{G}^{c} = \sum_{p,q}\mathbf{s}(p,q) \mathbb{E}^{c}_{p,q}
\label{eq:paste_accumulate}
\end{equation}

where $c\in[0,C)$, $i\in[0,2H)$, $j\in[0,2W)$, $p\in[-H/2,H/2]$, $q\in[-W/2,W/2]$, $\mathbf{s}_{p,q}$ is the similarity score of $(p, q)$-shifting, and $\mathbb{E}_{p,q}$ is the projection of $\mathbb{F}$’s $(p,q)$-shifted copy on the $(2H, 2W)$ grid. Namely, $\mathbb{E}^{c}_{p,q}(i+p+H/2,j+q+W/2)= \mathbb{F}^{c}(i,j)$, with $\mathbb{E} \in \mathbf{R}^{C\times 2H\times 2W}$, $\mathbb{F} \in \mathbf{R}^{C\times H\times W}$, $i\in[0,H)$ and $j\in[0,W)$.

We compute the similar score $\mathbf{s}(p,q)$ of current $(p,q)$-shifting using the overlapping region based on the following equation:

\begin{equation}
    \mathbf{s}(p,q) = -\frac{\sum_{m, n, c} \|\mathbb{F}^{c}(m, n) - \mathbb{F}^{c}(m-p, n-q)\|^2}{\sum_{m, n, c} \|\mathbb{F}^{c}(m, n)\|^2}
\label{eq:selfsim_per_shift}    
\end{equation}

Here, $m\in[max(0, p), min(p+H,H)]$ and $n\in[max(0,q), min(q+W,W)]$ indicate the overlapping region between current $(p, q)$-shifted copy and the original copy. $\|\mathbb{F}\|_2$ is the L2 norm of $\mathbb{F}$. The dominator $\sum_{m, n, c} \|\mathbb{F}_{m, n}^{c}\|^2$ is used for denormalization such that the scale of $\mathbf{s}(p,q)$ is independent of the scale of $\mathbb{F}$. Figure \ref{fig:framework}(a) shows how the self-similarity score is computed at shifting $(-H/2,-W/2)$. Note that self-similarity map is not symmetric with respect to its center as the dominator of Equation~\ref{eq:selfsim_per_shift} is not symmetric with respect to the center. Full animation of computing the self-similarity map can be found in the supplementary video. We will apply some simple transformation on $\mathbf{s}$ before using it in Equation~\ref{eq:paste_accumulate}, specifically one convolutional layer and one activation layer in our implementation. 

\begin{figure}
\centering
    \includegraphics[width=0.11\textwidth]{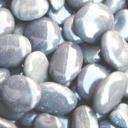}
    \includegraphics[width=0.11\textwidth,trim=80 20 80 20,clip]{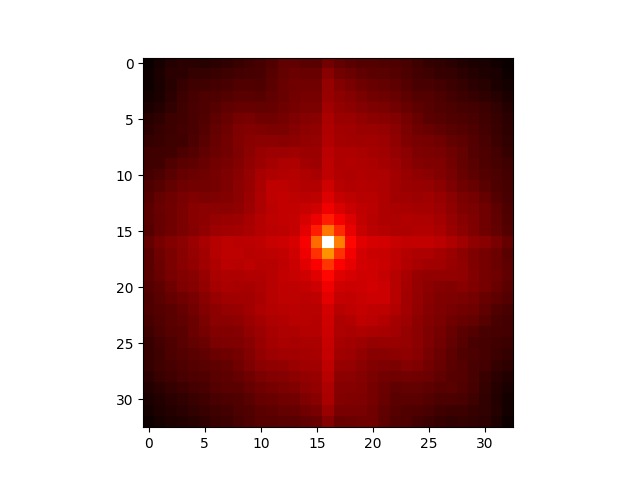}
    \includegraphics[width=0.11\textwidth,trim=80 20 80 20,clip]{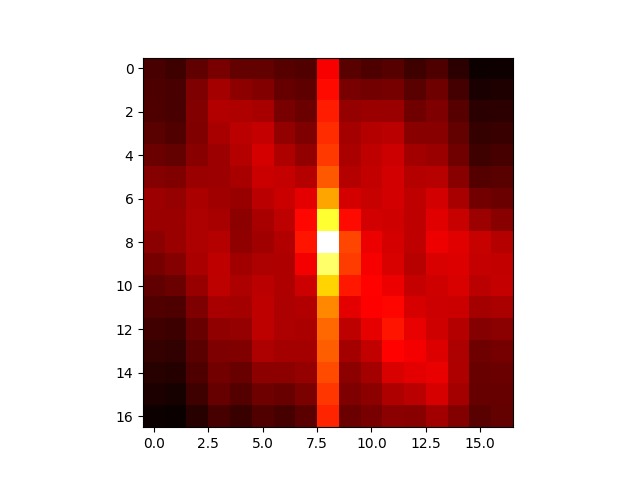}
    \includegraphics[width=0.11\textwidth,trim=80 20 80 20,clip]{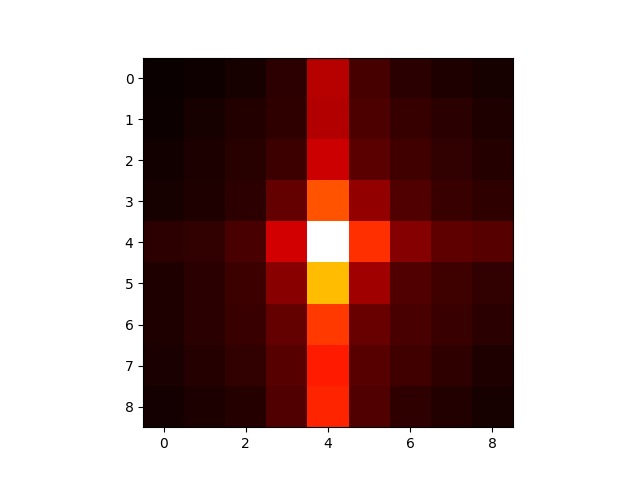}
    \includegraphics[width=0.11\textwidth]{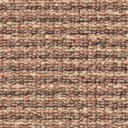}
    \includegraphics[width=0.11\textwidth,trim=80 20 80 20,clip]{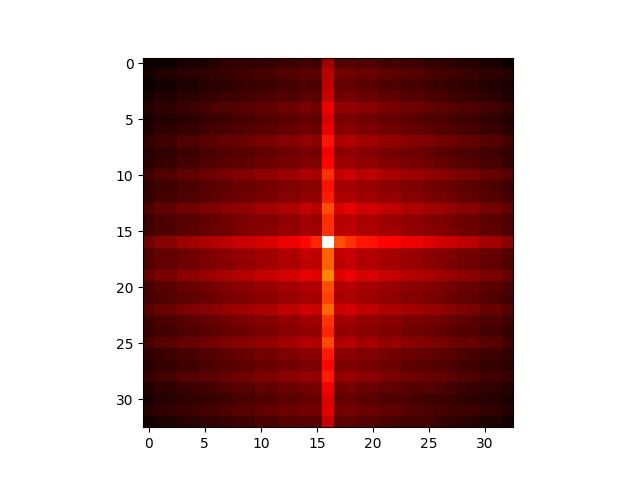}
    \includegraphics[width=0.11\textwidth,trim=80 20 80 20,clip]{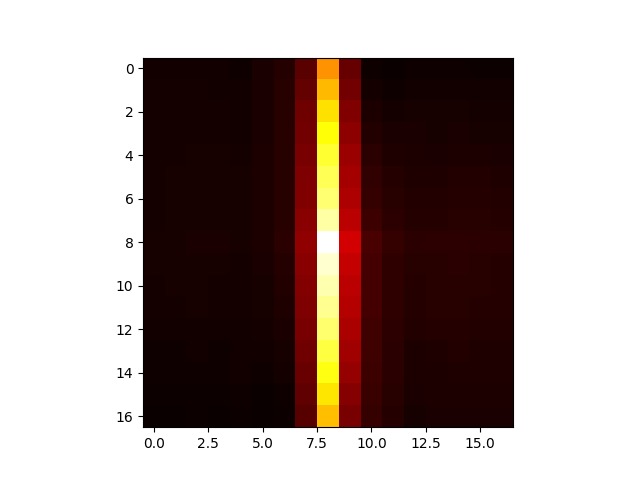}
    \includegraphics[width=0.11\textwidth,trim=80 20 80 20,clip]{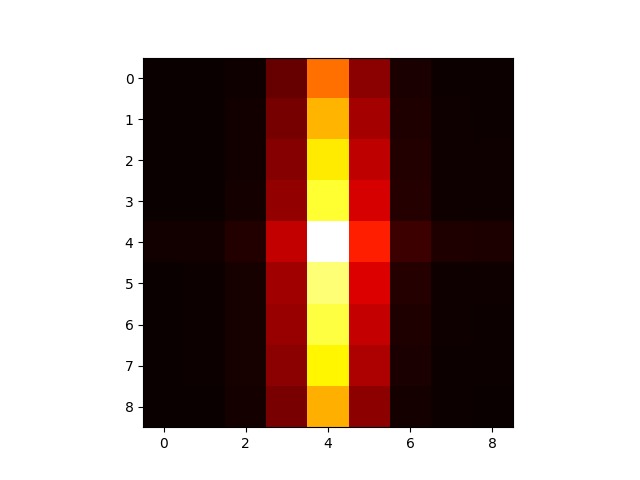} \\
    \includegraphics[width=0.11\textwidth]{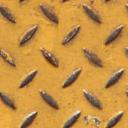}
    \includegraphics[width=0.11\textwidth,trim=80 20 80 20,clip]{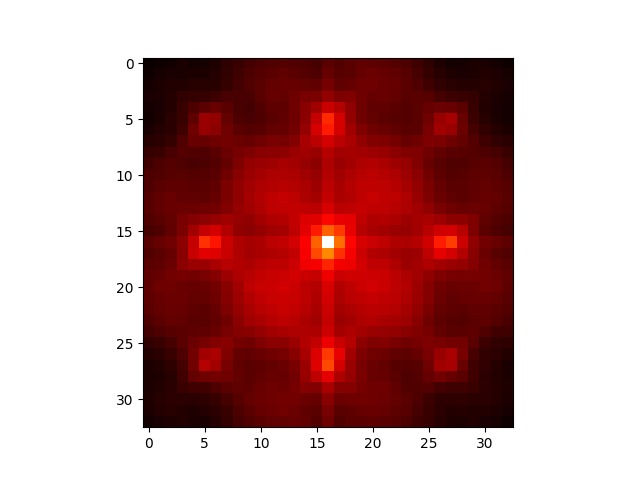}
    \includegraphics[width=0.11\textwidth,trim=80 20 80 20,clip]{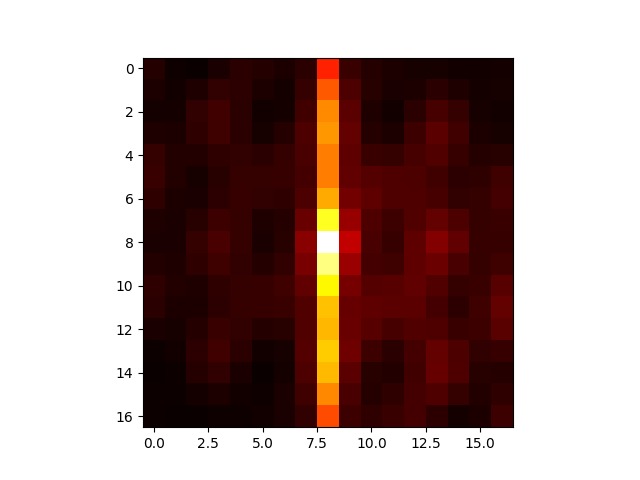}
    \includegraphics[width=0.11\textwidth,trim=80 20 80 20,clip]{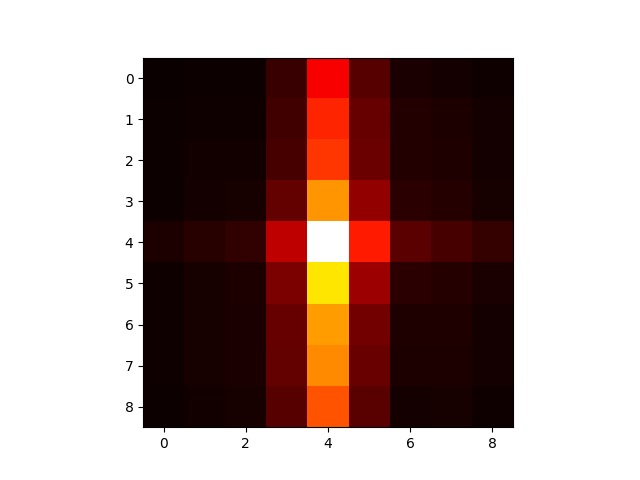}
    \includegraphics[width=0.11\textwidth]{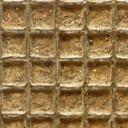}
    \includegraphics[width=0.11\textwidth,trim=80 20 80 20,clip]{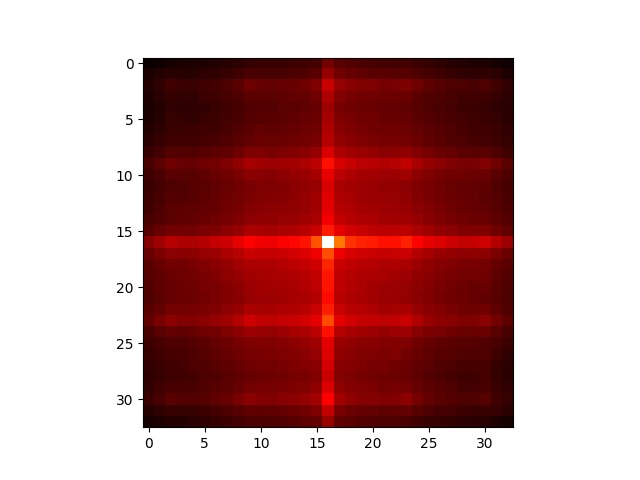}
    \includegraphics[width=0.11\textwidth,trim=80 20 80 20,clip]{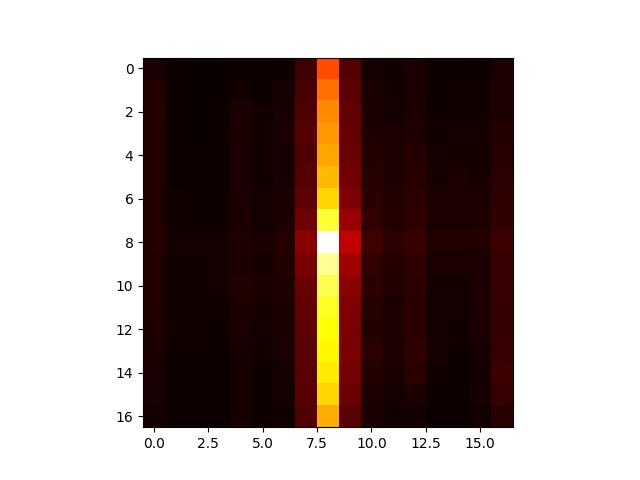}
    \includegraphics[width=0.11\textwidth,trim=80 20 80 20,clip]{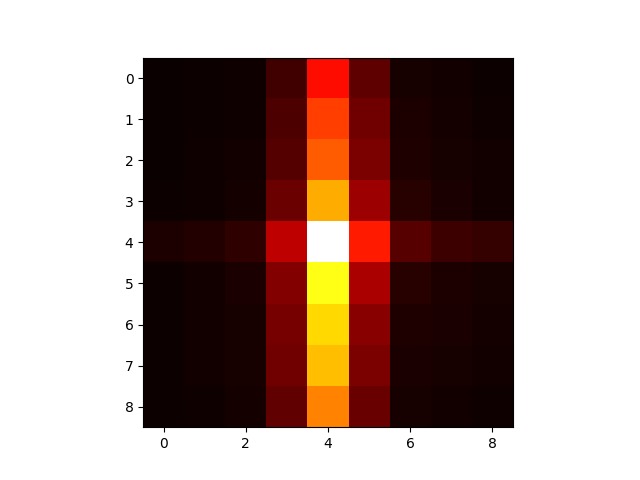} 
    \caption{Self-similarity Maps. We show the input texture images and the visualization of their self-similarity maps at 3 different scales ($1/4\times 1/4, 1/8\times 1/8, 1/16\times 1/16$). The first texture image exhibits more obvious self-similarity patterns at the second scale, while other three texture images exhibits more obvious self-similarity patterns at the first scale.}
\label{fig:simmapsample}
\vspace{-0.2cm}
\end{figure}

As shown in Equation \ref{eq:selfsim_per_shift}, the similarity score for a shift of $(p,q)$ along the width and height axis, respectively, is calculated as the L2 distance between the \textit{un-shifted} and \textit{shifted} copies of the feature map, normalized by the norm of the un-shifted copy's overlapping region.
%with the spatial size of their non-zero overlap and feature norm
%Recall, Equations \ref{eq:paste_accumulate} and \ref{eq:selfsim_per_shift} work on the $(p,q)$-shifting of $\mathbb{F}$ to simplify the equations. 
So, a shift of $(p=0,q=0)$ gives the maximum score because there is no shifting and it exactly matches the original copy. Computing self-similarity maps can be efficiently implemented with the help of existing convolution operations. Details are discussed in the supplementary file. 

%Note that computing self-similarity maps in parallel can be done with the help of existing convolution operations implemented in DL libraries. 
%Some examples of the similarity map visualization is shown in Figure \ref{fig:simmapsample}. 
We compute the self-similarity maps at multiple scales. Different texture images may exhibit more obvious self-similarity patterns on a specific scale than other scales, as shown in Figure~\ref{fig:simmapsample}. 
%For instance, in Figure~\ref{fig:simmapsample}, the first texture image shows more obvious self-similarity pattern at second scale while other texture images show more obvious self-similarity pattern at the first scale. 

\textbf{Feature (Texture) Expansion via Transposed Convolution Operation}:
Note that the process of pasting shifted feature maps and aggregating them to create larger feature maps is equivalent to the operation of a standard transposed convolution in deep neural networks. For the given filter and input data, a transposed convolution operation simply copies the filter weighted by the respective center entry's data value in the input data into a larger feature output grid, and perform a summation. In fact, our proposed Equation \ref{eq:paste_accumulate} is similar with a transposed convolution. Specifically, we apply transposed convolutions with a stride of $1$, treating the feature map $\mathbb{F} \in \mathbf{R}^{C\times H\times W}$ as the transposed convolution filter, and the similarity map $\mathbb{S} \in \mathbf{R}^{1\times (H+1) \times (W+1)}$, given by Equation \ref{eq:selfsim_per_shift}, as the input to the transposed convolution. This results in an output feature map $\mathbb{G}$ of size $C \times 2H \times 2W$. Figure~\ref{fig:framework}(b) shows how the transposed convolution is done using the encoded input texture as filters and the first entry in the self-similarity map as input. Full animation of the transposed convolution operation can be found in the supplementary video. 

\subsection{Architecture}

Figure \ref{fig:framework}(c) illustrates our overall texture synthesis framework. It relies on a UNet-like architecture. The encoder extracts deep features of the input texture patch at several scales. We then apply our proposed transposed convolution-based feature map expansion technique at each scale. The resulting expanded feature map is then passed onto a standard decoder layer.
%(bilinear upscale plus normal convolution) to further upscale the feature map, which will then be combined with the corresponding feature map at the next level, and so on. 
Our network is fully differentiable, allowing us to train our model with stochastic gradient-based optimizations in an end-to-end manner. The four main components of our framework in Figure \ref{fig:framework}(c) are:

% Our overall framework is shown in Figure~\ref{fig:framework}. We will now discuss the 4 main components of our network:
\begin{enumerate}
    \item \textbf{Encoder}: Learns to encode the input texture image into deep features at different scales or levels.
    \item \textbf{Self-Similarity Map Generation}: Constructs guidance maps from the encoded features to weight the shifted feature maps in the shift, paste and aggregate process of feature map expansion (see Equation \ref{eq:selfsim_per_shift} and Figure~\ref{fig:framework}(a)).
    \item \textbf{Transposed Convolution Operation}: Applies spatially varying transposed convolution operations, treating the encoded feature maps directly as \textit{filters} and the self-similarity maps as \textit{inputs} to produce expanded feature maps, as shown in Figure~\ref{fig:framework}(b). Note that, unlike traditional transposed convolution layers, ours transposed convolution filters are not learning parameters. More details about the difference between our transposed convolution operation and traditional transposed convolution layer can be found in the suppmental file.
    \item \textbf{Decoder}: Given the already expanded features from the transposed convolution operations at different scales, we follow the traditional decoder network design that uses standard convolutional layers followed by bilinear upsampling layers to aggregate features at different scales, and generate the final output texture, as shown in the last row of Figure~\ref{fig:framework}(c).
\end{enumerate}

As described above, our proposed texture expansion technique is performed at multiple feature representation levels, allowing us to capture both diverse features and their optimal aggregation weights. Unlike previous approaches that rely on heuristics or graph-base techniques to identify the optimal overlap of shifted textures, our approach formulates the problem as a direct generation of larger texture images conditioned on optimally assembled deep features at multiple scales. This makes our approach desirable as it is data-driven and generalizable for various textures.

\subsection{Loss Functions}
\label{sec:loss}

\begin{figure*}
\centering
\scalebox{0.8}{
\begin{tabular}{c|ccc|c}
% \multicolumn{1}{c}{} & \multicolumn{4}{c}{} \\
input & no perceptual loss & no style loss & non GAN loss & full loss \\
\includegraphics[width=0.1\textwidth]{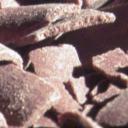}&
\includegraphics[width=0.2\textwidth]{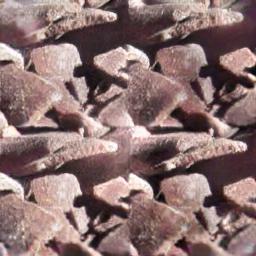}&
\includegraphics[width=0.2\textwidth]{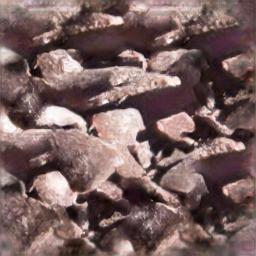}&
\includegraphics[width=0.2\textwidth]{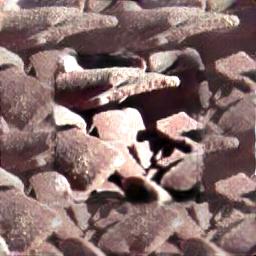}&
\includegraphics[width=0.2\textwidth]{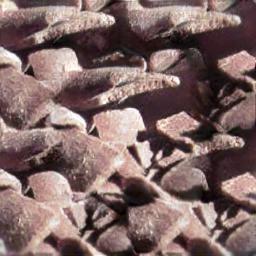}\\
\end{tabular}
}
\caption{Ablation study on the components of loss functions.}
\vspace{-0.2cm}
\label{fig:loss_ablate}
\end{figure*}

During training, given a random image with size (2$H$, 2$W$), denoted as $I_{target}$, its center crop image with size ($H$,$W$) will be the input to the network, denoted as $I_{input}$. We train the network to predict an output image $I_{out}$ with the size (2$H$, 2$W$). Both VGG-based perceptual loss, style loss and GAN loss are used to train the network. The perceptual loss and style loss are defined between $I_{target}$ and  $I_{out}$ at the full resolution of $(2H, 2W)$; meanwhile, the GAN loss is defined on the random crops at the resolution of $(H, W)$. Details are discussed below.

\textbf{VGG-based perceptual loss and style loss}. Perceptual loss and style loss are defined following Gatys et al.~\cite{gatys2015neural}. 

The perceptual loss and style loss are defined as:

\begin{equation}
    \mathcal{L}_{perceptual} = \sum_{p=0}^{P-1} {\frac{\|{\Psi}_{p}^{\mathbf{I}_{out}}-{\Psi}_{p}^{\mathbf{I}_{target}}\|_{1}}{N_{{\Psi}_{p}^{\mathbf{I}_{target}}}}}
\end{equation}

% \begin{equation}
% \begin{split}
%  \mathcal{L}_{style} = &\sum_{p=0}^{P-1}{\frac{1}{C_{p} C_{p}} {\Big|\Big|K_p\big({{\big(\Psi}_{p}^{\mathbf{I}_{out}}\big)}^{\intercal}{{\big(\Psi}_{p}^{\textbf{I}_{out}}\big)} - \\ &{\big({\Psi}_{p}^{\mathbf{I}_{target}}\big)}^{\intercal}{\big({\Psi}_{p}^{\mathbf{I}_{target}}\big)\big)}\Big|\Big|}_{1} }
% \end{split}
% \label{eq:style_loss}
% \end{equation}

\begin{equation}
 \mathcal{L}_{style_{out}} = \sum_{p=0}^{P-1}{\frac{1}{C_{p} C_{p}} {\Big|\Big|K_p\big({{\big(\Psi}_{p}^{\mathbf{I}_{out}}\big)}^{\intercal}{{\big(\Psi}_{p}^{\textbf{I}_{out}}\big)} - {\big({\Psi}_{p}^{\mathbf{I}_{target}}\big)}^{\intercal}{\big({\Psi}_{p}^{\mathbf{I}_{target}}\big)\big)}\Big|\Big|}_{1} } 
\label{eq:style_loss}
 \end{equation}

% \begin{equation}
% \begin{aligned}
%  \mathcal{L}_{style_{out}} = &\sum_{p=0}^{P-1}{\frac{1}{C_{p} C_{p}} {\Big|\Big|K_p\big({{\big(\Psi}_{p}^{\mathbf{I}_{out}}\big)}^{\intercal}{{\big(\Psi}_{p}^{\textbf{I}_{out}}\big)} - \\ 
%  &{\big({\Psi}_{p}^{\mathbf{I}_{target}}\big)}^{\intercal}{\big({\Psi}_{p}^{\mathbf{I}_{target}}\big)\big)}\Big|\Big|}_{1} } \\
%  \end{aligned}
% \label{eq:style_loss}
%  \end{equation}

Here, $N_{{\Psi}_{p}^{\mathbf{I}_{target}}}$ is the number of entries in ${\Psi}_{p}^{\mathbf{I}_{target}}$. The perceptual loss computes the $L^1$ distances between both $\mathbf{I}_{out}$ and $\mathbf{I}_{target}$, but after projecting these images into higher level feature spaces using an ImageNet-pretrained VGG-19~\cite{simonyan2014very}. ${\Psi}_{p}^{\mathbf{I}_{*}}$ is the activation map of the $p$th selected layer given original input $\mathbf{I}_{*}$. We use feature from $2$-nd, $7$-th, $12$-th, $21$-st and $30$-th layers corresponding to the output of the ReLU layers at each scale. In Equation (4), the matrix operations assume that the high level features $\Psi(x)_p$ is of shape $(H_pW_p)\times C_p$, resulting in a $C_p\times C_p$ Gram matrix, and $K_p$ is the normalization factor $1/C_pH_pW_p$ for the $p$th selected layer.

% \textbf{VGG-based style loss}. The style loss is defined based on the Gram matrix of these features. 

% \begin{equation}
% \begin{split}
%  \mathcal{L}_{style} = &\sum_{p=0}^{P-1}{\frac{1}{C_{p} C_{p}} {\Big|\Big|K_p\big({{\big(\Psi}_{p}^{\mathbf{I}_{out}}\big)}^{\intercal}{{\big(\Psi}_{p}^{\textbf{I}_{out}}\big)} - \\ &{\big({\Psi}_{p}^{\mathbf{I}_{target}}\big)}^{\intercal}{\big({\Psi}_{p}^{\mathbf{I}_{target}}\big)\big)}\Big|\Big|}_{1} }
% \end{split}
% \end{equation}

% Where the matrix operations assume that the high level features $\Psi(x)_p$ is of shape $(H_pW_p)\times C_p$, resulting in a $C_p\times C_p$ Gram matrix, and $K_p$ is the normalization factor $1/C_pH_pW_p$ for the $p$th selected layer.

\textbf{GAN loss}. 
The discriminator takes the concatenation of $I_{input}$ and a random crop of size $(H, W)$ from either $I_{out}$ or $I_{target}$ as input. Denote the random crop from $I_{out}$ as $I_{randcrop}^{out}$ and the random crop from $I_{target}$ as $I_{randcrop}^{target}$. The intuition of using concatenation is to let the discriminator learn to classify whether $I_{input}$ and $I_{randcrop}^{*}$ is a pair of two similar texture patches or not. 
We randomly crop 10 times for both $I_{out}$ and $I_{target}$ and sum up the losses. 

%The discriminator is trained to classify $<I_{input}, I_{randcrop}^{target}>$ as true and $<I_{input}, I_{randcrop}^{out}>$ as false; while the generator is trained so that discriminator can classify $<I_{input}, I_{randcrop}^{out}>$ as true. We randomly crop 10 times for both $I_{input}$ and $I_{target}$ and sum them up. 

%\textbf{Loss function ablation study}. 
\textbf{Ablation Study}. These 3 losses are summed up with the weights of 0.05, 120 and 0.2 respectively. We find that all of them are useful and necessary. As shown in Figure~\ref{fig:loss_ablate}, without perceptual loss, the result just looks like the naive tiling of the inputs; no style loss makes the border region blurry; and no GAN loss leads to obvious discrepancy between the center region and the border region.

%% file: sec_comp.tex
\section{Experiments and Comparisons}

% comparison images
\begin{figure*}
\centering
\scalebox{0.68}{
\addtolength{\tabcolsep}{-3pt}
\begin{tabular}{c|ccccccccc}
% \addtolength{\tabcolsep}{-3pt}
% \setlength{\tabcolsep}{1em}
\multicolumn{1}{c}{} & \multicolumn{9}{c}{} \\
input & transposer(ours) & self-tuning & pix2pixHD & SinGAN$^{(*)}$ & Non-stat.$^{(*)}$ & WCT & DeepText.$^{*}$ & Text. Mixer & ground truth \\
\hline
\hline
\includegraphics[width=0.07\textwidth]{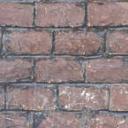} &
\includegraphics[width=0.14\textwidth]{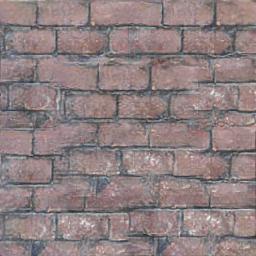} &
\includegraphics[width=0.14\textwidth]{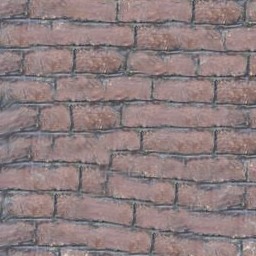} &
\includegraphics[width=0.14\textwidth]{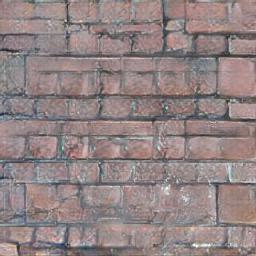} & 
\includegraphics[width=0.14\textwidth]{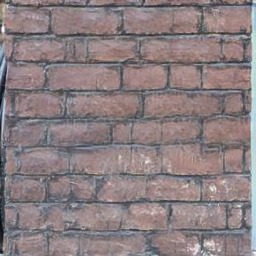} &
\includegraphics[width=0.14\textwidth]{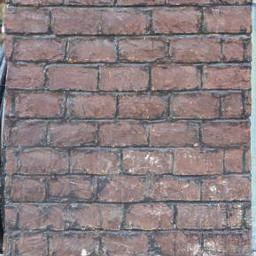} &
\includegraphics[width=0.14\textwidth]{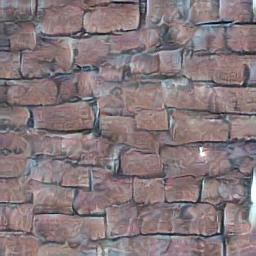}&
\includegraphics[width=0.14\textwidth]{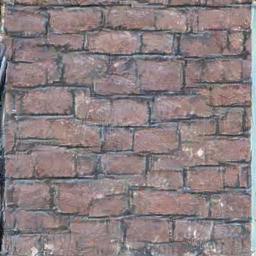} &
\includegraphics[width=0.14\textwidth]{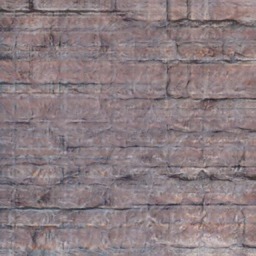} &
\includegraphics[width=0.14\textwidth]{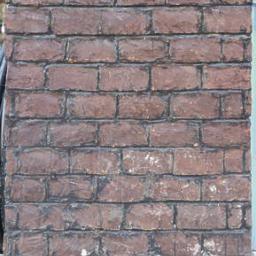} \\
\includegraphics[width=0.07\textwidth]{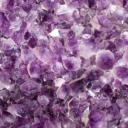} &
\includegraphics[width=0.14\textwidth]{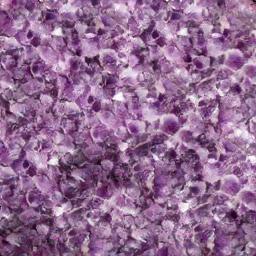} &
\includegraphics[width=0.14\textwidth]{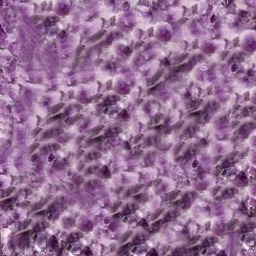} &
\includegraphics[width=0.14\textwidth]{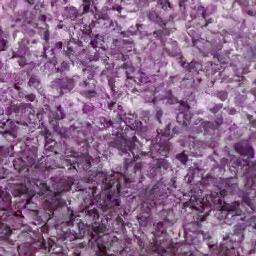} &
\includegraphics[width=0.14\textwidth]{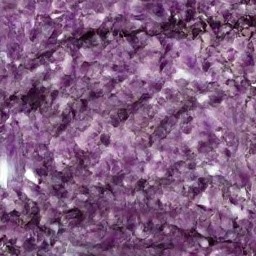} &
\includegraphics[width=0.14\textwidth]{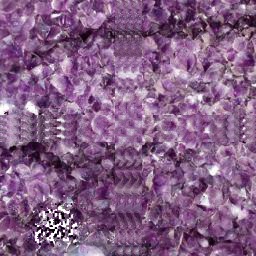} &
\includegraphics[width=0.14\textwidth]{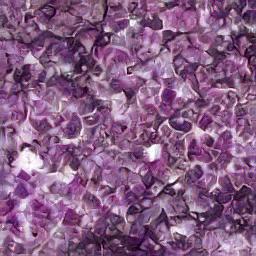} &
\includegraphics[width=0.14\textwidth]{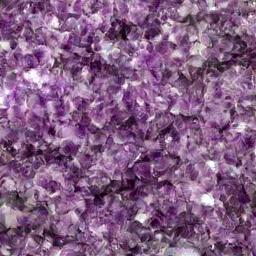} &
\includegraphics[width=0.14\textwidth]{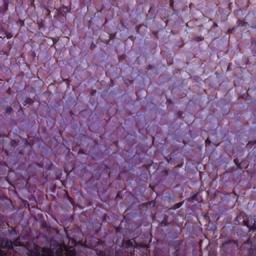} &
\includegraphics[width=0.14\textwidth]{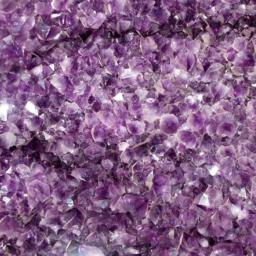} \\
\hline
\hline
\includegraphics[width=0.07\textwidth]{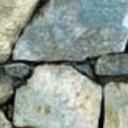} &
\includegraphics[width=0.14\textwidth]{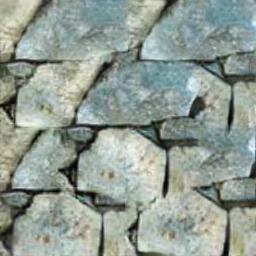} &
\includegraphics[width=0.14\textwidth]{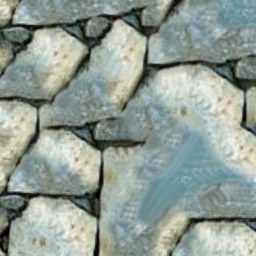} &
\includegraphics[width=0.14\textwidth]{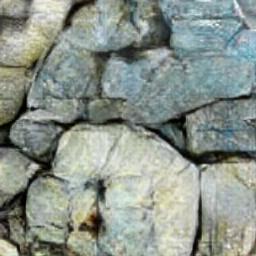} &
\includegraphics[width=0.14\textwidth]{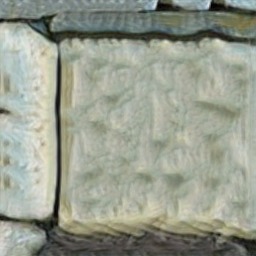} &
\includegraphics[width=0.14\textwidth]{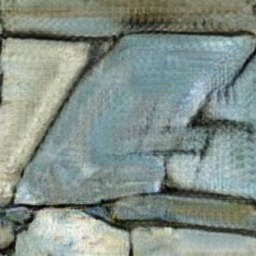} &
\includegraphics[width=0.14\textwidth]{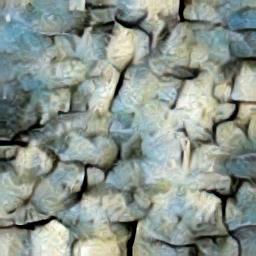} &
\includegraphics[width=0.14\textwidth]{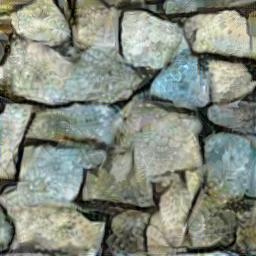} &
\includegraphics[width=0.14\textwidth]{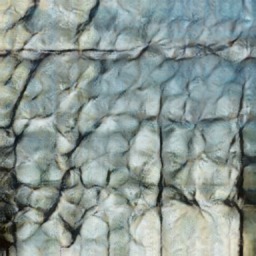} &
\includegraphics[width=0.14\textwidth]{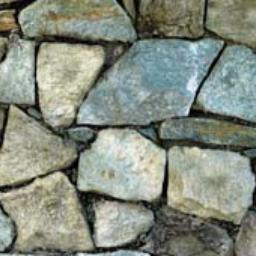} \\

\includegraphics[width=0.07\textwidth]{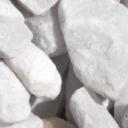} &
\includegraphics[width=0.14\textwidth]{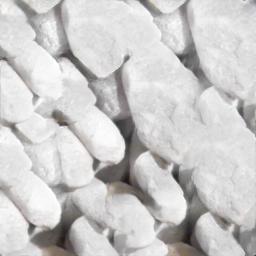} &
\includegraphics[width=0.14\textwidth]{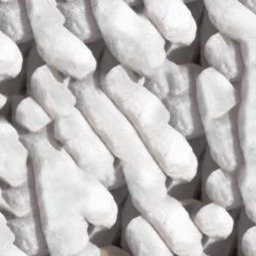} &
\includegraphics[width=0.14\textwidth]{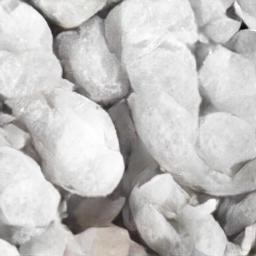} &
\includegraphics[width=0.14\textwidth]{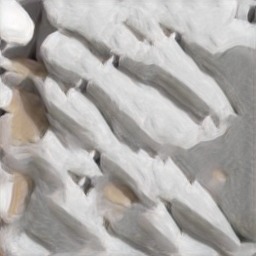} &
\includegraphics[width=0.14\textwidth]{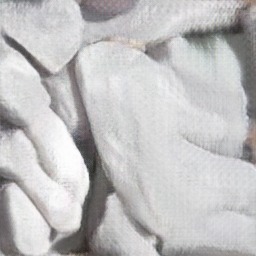} &
\includegraphics[width=0.14\textwidth]{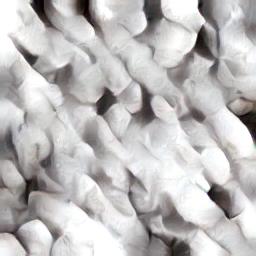} &
\includegraphics[width=0.14\textwidth]{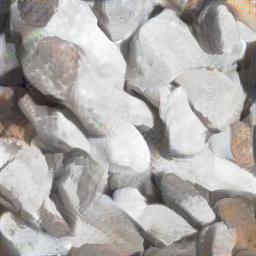} &
\includegraphics[width=0.14\textwidth]{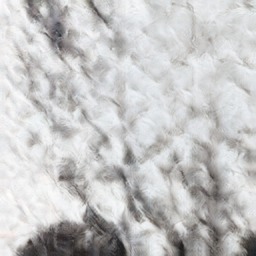} &
\includegraphics[width=0.14\textwidth]{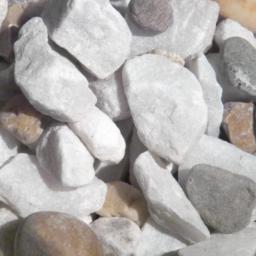} \\
\end{tabular}
% \addtolength{\tabcolsep}{-4pt}   
}
\caption{Results of different approaches on $128\times 128$ to $256\times 256$ texture synthesis. For SinGAN($^{*}$) and Non-stat.$^{(*)}$ results, the first two rows show the results when training with direct access to exact-size ground truth; the remaining 2 rows show the results without them accessing the ground truth. In this paper, unless specified, our results (transposer) uses self-similarity map as transposed convolution inputs by default.}
\label{fig:result256}
\vspace{-0.2cm}
\end{figure*}
\subsection{Dataset \& Training}

To train our network, we collected a large dataset of texture images. We downloaded 55,583 images from 15 different texture image sources \cite{cimpoi14describing,sharan2014flickr,dai2014synthesizability,burghouts2009material,outex,picard2010vistex,abdelmounaime2013new,fritz2004kth,mallikarjuna22006kth2}. The total dataset consists of texture images with a wide variety of patterns, scales, and resolutions. We randomly split the dataset to create a training set of 49,583 images, a validation set of 1,000 images, and a test set of 5,000 images. All generation results and evaluation results in the paper are from the test set. When using these images, we resize them to the target output size as the ground truth and the center cropping of it as input. 

Our network utilizing the transposed convolution operation is implemented using the existing PyTorch interface without custom CUDA kernels. We trained our model on 4 DGX-1 stations with 32 total NVIDIA Tesla V100 GPUs using synchronized batch normalization layers~\cite{ioffe2015batch}. For 128$\times$128 to 256$\times$256 synthesis, we use batch size 8 and trained for 600 epochs. The learning rate is set to be 0.0032 at the beginning and decreased by $1/10$ every 150 epochs. For 256$\times$256 to 512$\times$512 synthesis, we fine-tuned the model based on the pre-trained one for 128$\times$128 to 256$\times$256 synthesis for 200 epochs. While directly using 128 to 256 synthesis pre-trained model generates reasonable results,  fine-tuning leads to better quality.

\subsection{Baseline \& Evaluation Metrics}

\begin{table*}
    \centering
    \begin{tabular}{l|cc|cc}
    \multicolumn{1}{c}{} & \multicolumn{2}{c}{Time} & \multicolumn{2}{c}{Properties}  \\
    \hline
    % \multicolumn{1}{c}{} & \multicolumn{1}{|c}{256x256} & \multicolumn{1}{c}{512x512} \\
    Method & 256x256 & 512x512 & Generalizability & Size-increasing \\
    \hline
    Self-tuning\cite{kaspar2015self} & 140 s & 195 s & Good & Yes \\
    Non-stationary\cite{zhou2018non} & 362 mins & 380 mins & No & Yes \\
    SinGAN\cite{shaham2019singan} & 45 mins & 100 mins & No & Yes \\
    DeepTexture\cite{gatys2015texture} & 13 mins & 54 mins & No & No \\
    WCT\cite{yijun2017universal} & 7 s & 14 s & Medium & Yes \\
    pix2pixHD~\cite{wang2018high} & \textbf{11 ms} & \textbf{22 ms} & Medium & Yes \\
    Texture Mixer~\cite{yu2019texture} & - & 799 ms & Medium & Yes \\
    \textbf{transposer(ours)} & 43 ms & 260 ms & Good & Yes \\
    \end{tabular}
    \caption{Time required for synthesis at different spatial resolutions for various approaches and their corresponding properties. For Non-stationary and SinGAN, the reported time includes training time. All methods are run on one NVIDIA Tesla V100, except for Self-tuning which runs the default 8 threads in parallel on an Intel Core i7-6800K CPU @ 3.40GHz. }
    \label{tab:timing}
    \vspace{-0.2cm}
\end{table*}

\textbf{Baselines}. We compare against several baselines: 1) \textbf{Naive tiling} which simply tiles the input four times; 2) \textbf{Self-tuning} \cite{kaspar2015self}, the state-of-the-art optimization-based method; 3) \textbf{pix2pixHD} \cite{wang2018high}, the state-of-the-art image-to-image translation network where we add one more upsampling layer to generate an output 2x2 larger than the input; 4) \textbf{WCT}~\cite{yijun2017universal} is the style transfer method; 5) \textbf{DeepTexture}~\cite{gatys2015texture}, an optimization based using network features, for which we directly feed the ground truth as input; 6) \textbf{Texture Mixer}~\cite{yu2019texture}, a texture interpolation method where we set the interpolation source patches to be all from the input texture; 7) \textbf{Non-stationary (Non-stat.)}~\cite{zhou2018non} and \textbf{SinGAN}~\cite{shaham2019singan}, both of which overfit one model per texture. We train \textbf{Non-stat.} and \textbf{SinGAN} for two versions respectively; one version with direct access to the exact ground truth at the exact target size, and one version without access to target-size ground truth but only the input. In the paper, $^{*}$ will correspond to methods that either directly take ground truth images for processing or are overfitting the model to ground truth.

\begin{figure*}
\centering
\scalebox{0.85}{
\addtolength{\tabcolsep}{-3pt}   
\begin{tabular}{c|cccccccc}
\multicolumn{1}{c}{} & \multicolumn{8}{c}{} \\
input & transposer(ours) & self-tuning & pix2pixHD & SinGAN$^{*}$ & Non-stat.$^{*}$ & DeepTexure$^{*}$ & Text. Mixer & GroundTruth \\
\includegraphics[width=0.06\textwidth]{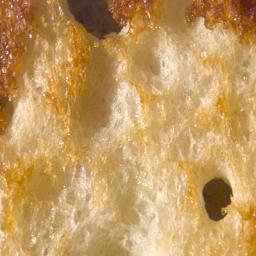} &
\includegraphics[width=0.12\textwidth]{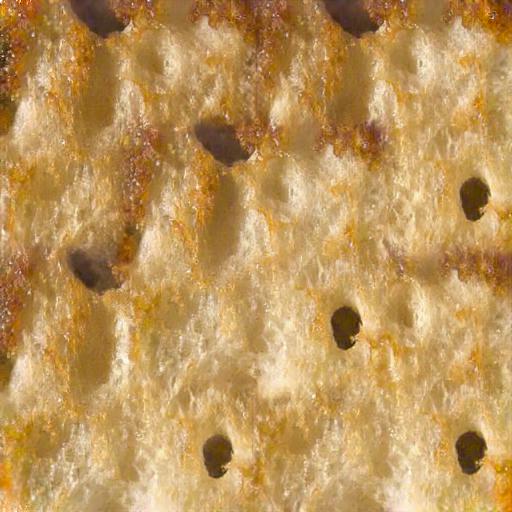} &
\includegraphics[width=0.12\textwidth]{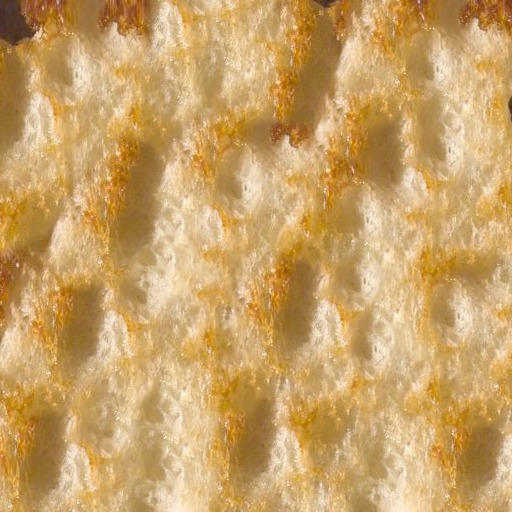} &
\includegraphics[width=0.12\textwidth]{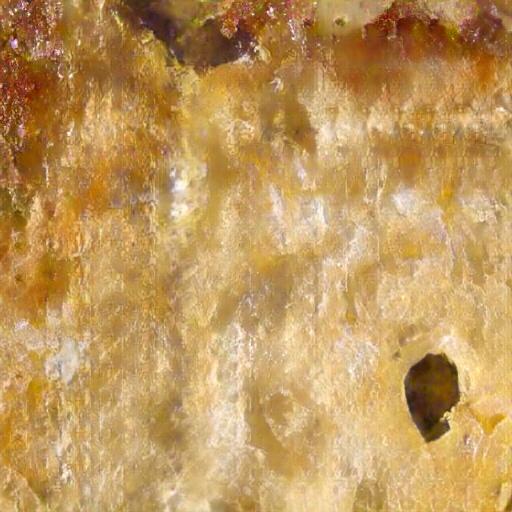} &
\includegraphics[width=0.12\textwidth]{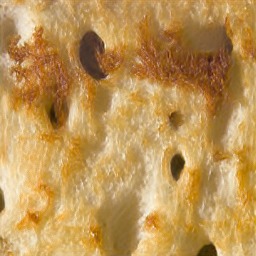} &
\includegraphics[width=0.12\textwidth]{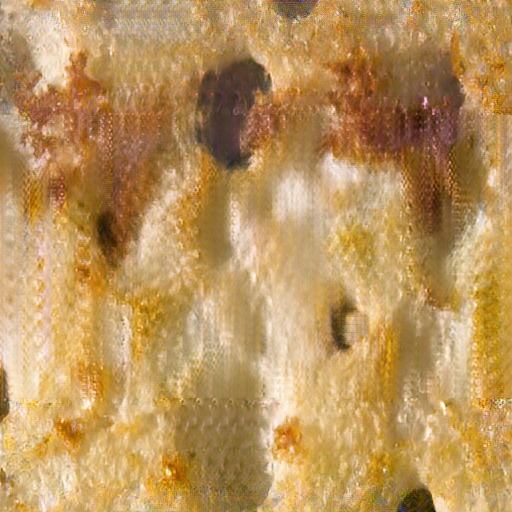} &
\includegraphics[width=0.12\textwidth]{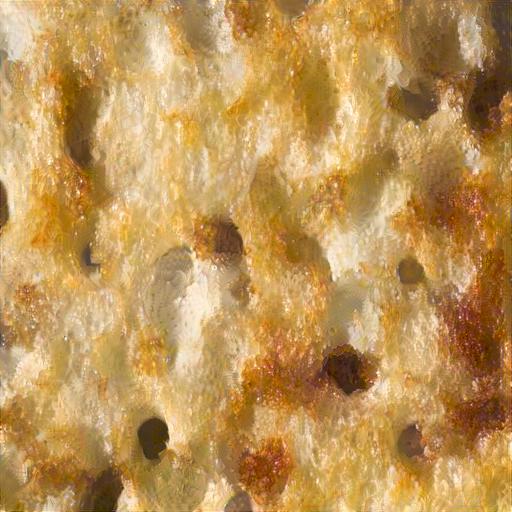} &
\includegraphics[width=0.12\textwidth]{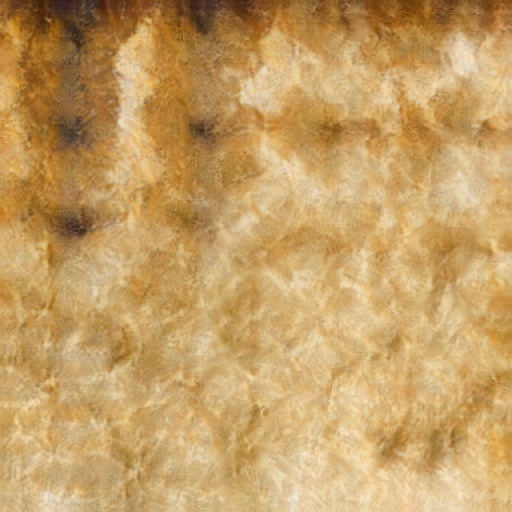} &
\includegraphics[width=0.12\textwidth]{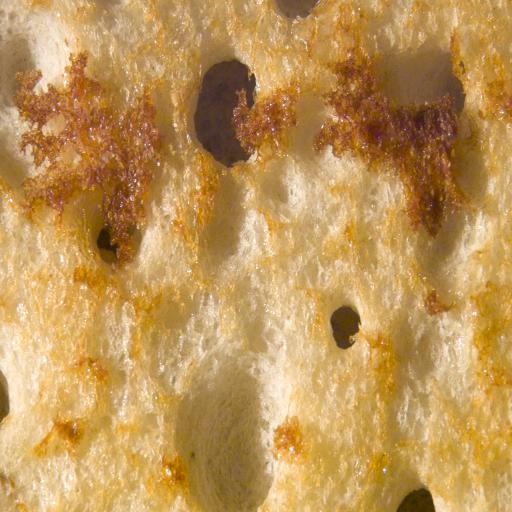} \\
\includegraphics[width=0.06\textwidth]{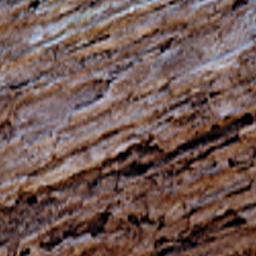} &
\includegraphics[width=0.12\textwidth]{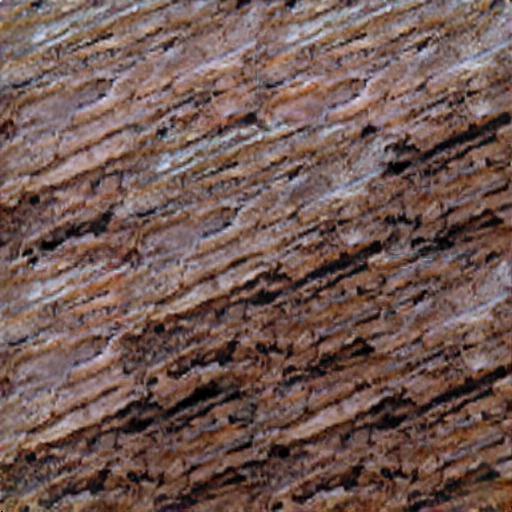} &
\includegraphics[width=0.12\textwidth]{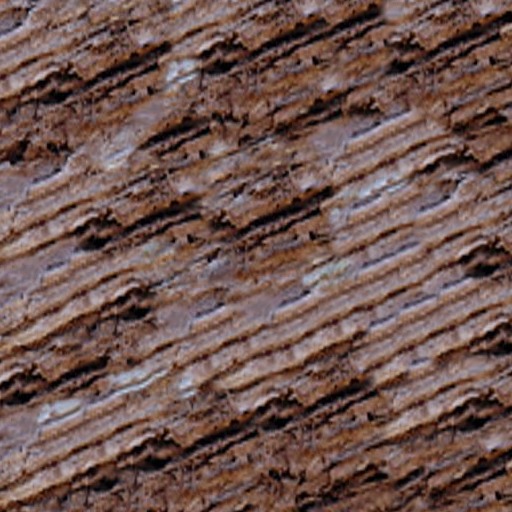} &
\includegraphics[width=0.12\textwidth]{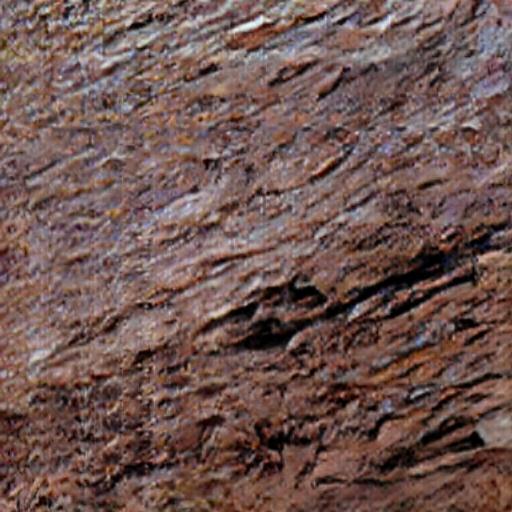} &
\includegraphics[width=0.12\textwidth]{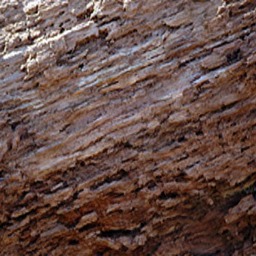} &
\includegraphics[width=0.12\textwidth]{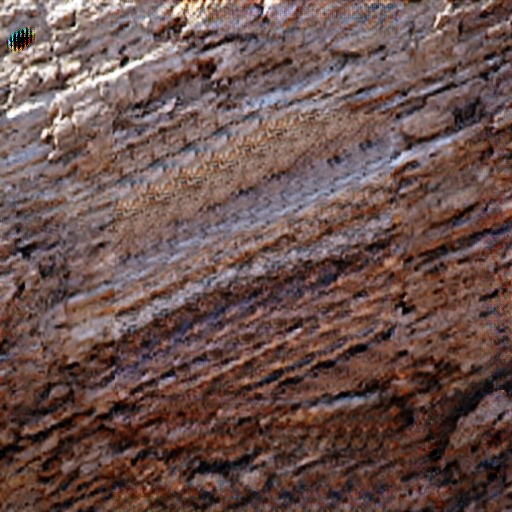} &
\includegraphics[width=0.12\textwidth]{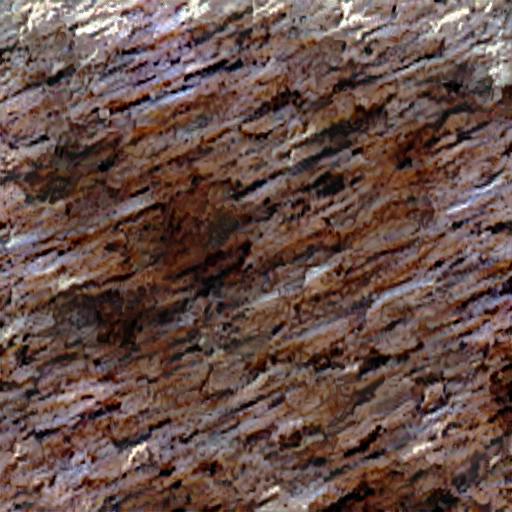} &
\includegraphics[width=0.12\textwidth]{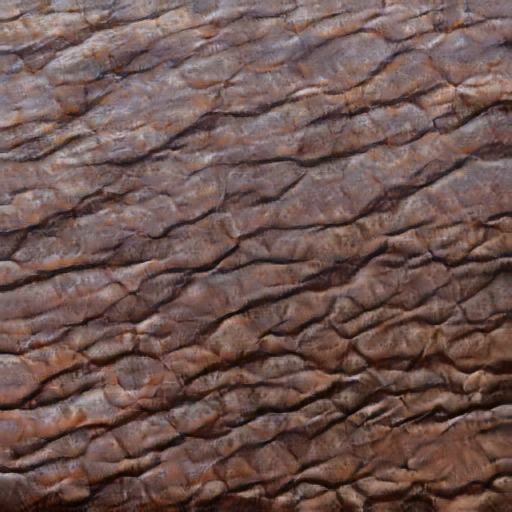} &
\includegraphics[width=0.12\textwidth]{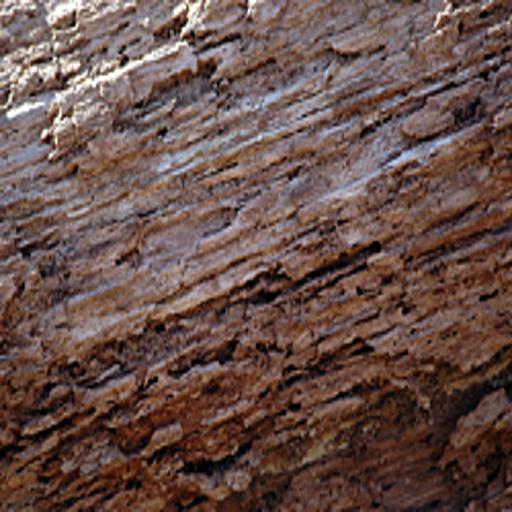} \\
\hline
\hline
\includegraphics[width=0.06\textwidth]{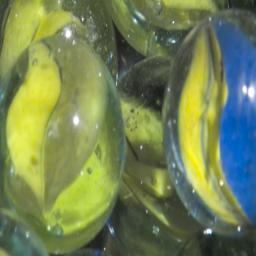} &
\includegraphics[width=0.12\textwidth]{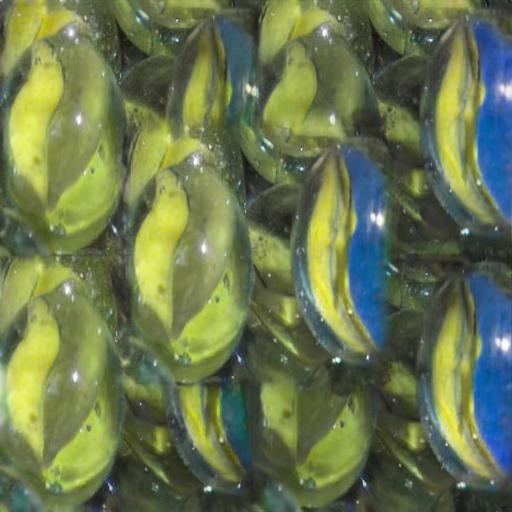} &
\includegraphics[width=0.12\textwidth]{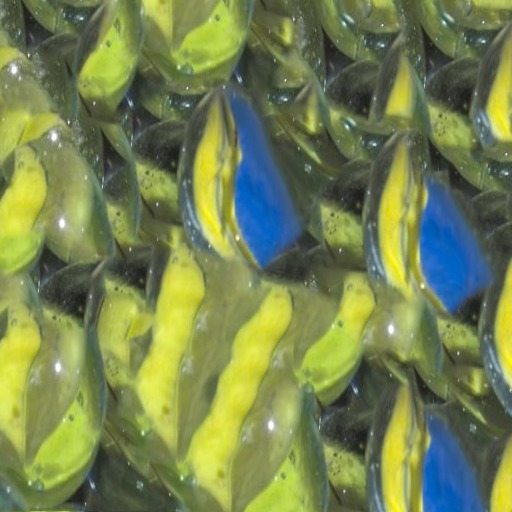} &
\includegraphics[width=0.12\textwidth]{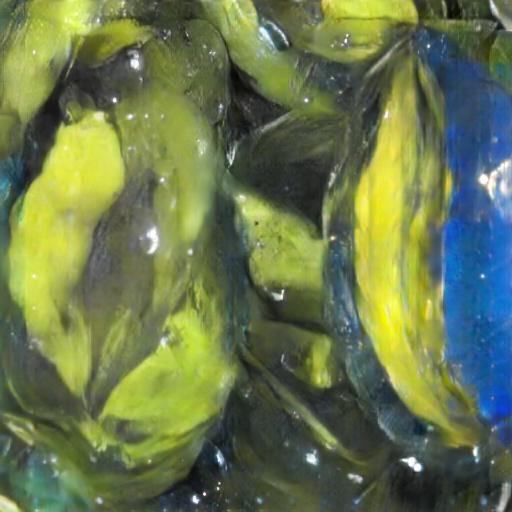} &
\includegraphics[width=0.12\textwidth]{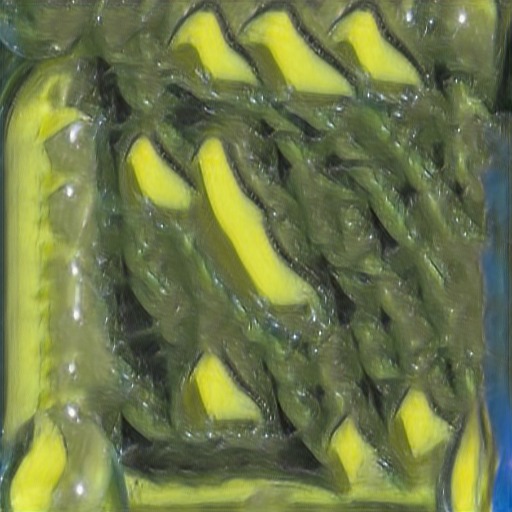} &
\includegraphics[width=0.12\textwidth]{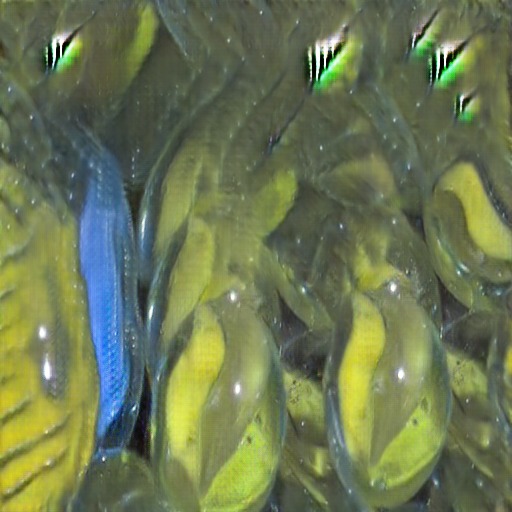} &
\includegraphics[width=0.12\textwidth]{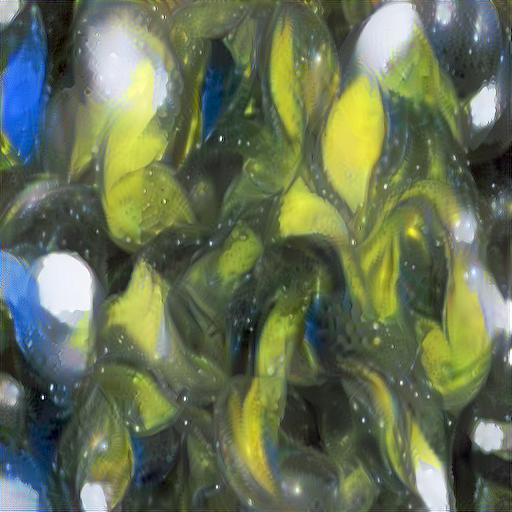} &
\includegraphics[width=0.12\textwidth]{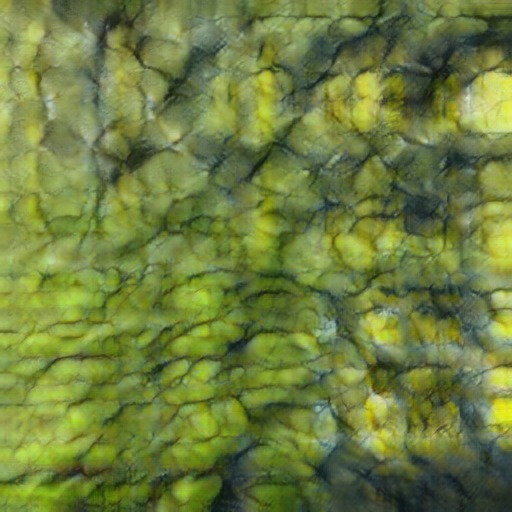} &
\includegraphics[width=0.12\textwidth]{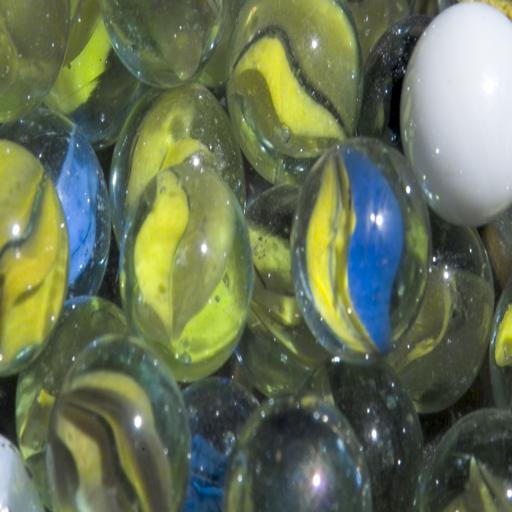} \\
\includegraphics[width=0.06\textwidth]{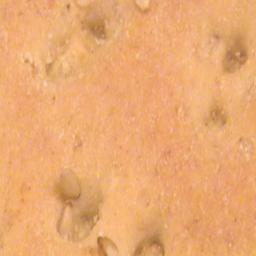} &
\includegraphics[width=0.12\textwidth]{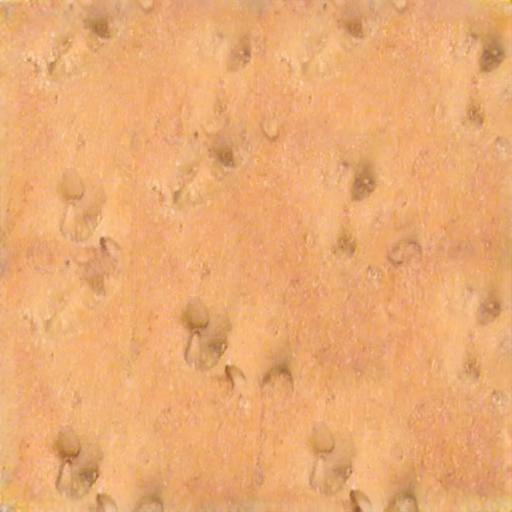} &
\includegraphics[width=0.12\textwidth]{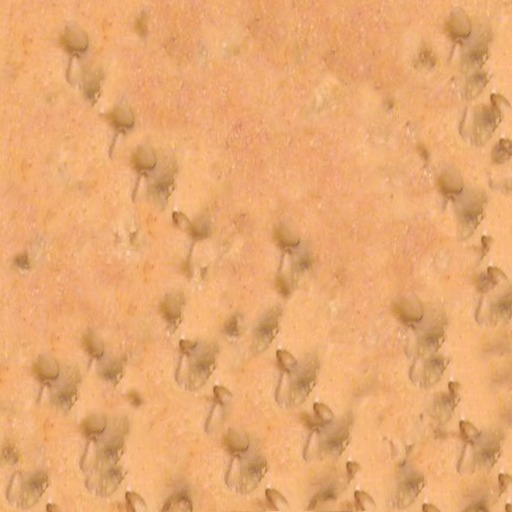} &
\includegraphics[width=0.12\textwidth]{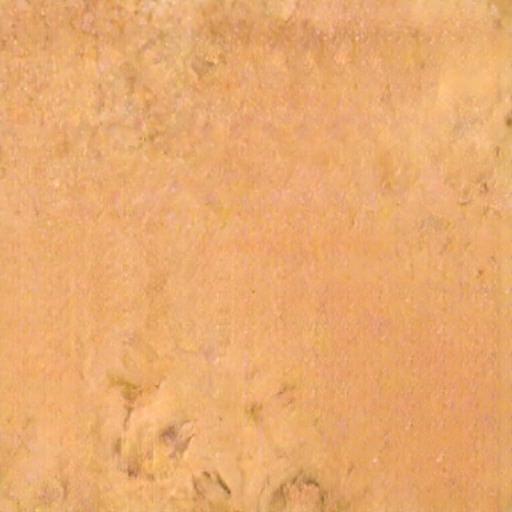} &
\includegraphics[width=0.12\textwidth]{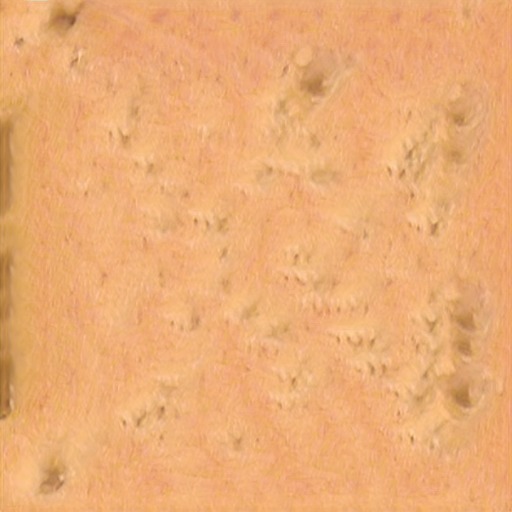} &
\includegraphics[width=0.12\textwidth]{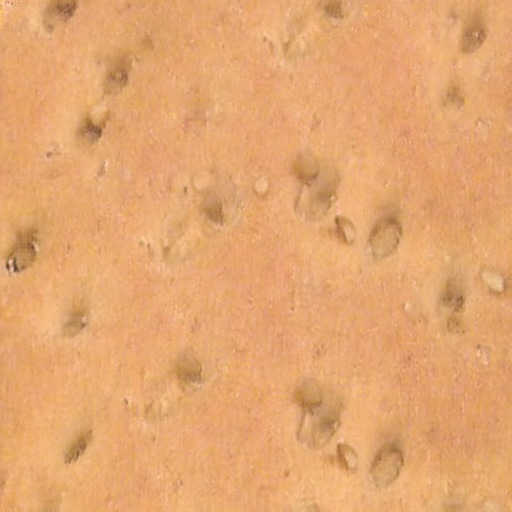} &
\includegraphics[width=0.12\textwidth]{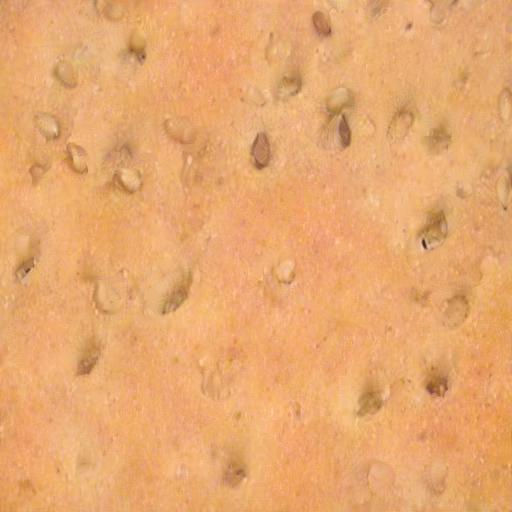} &
\includegraphics[width=0.12\textwidth]{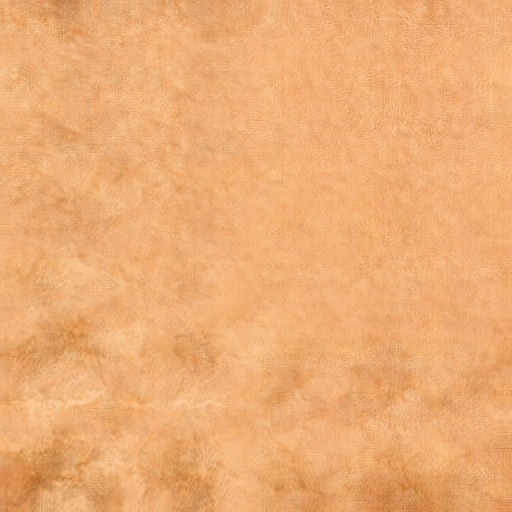} &
\includegraphics[width=0.12\textwidth]{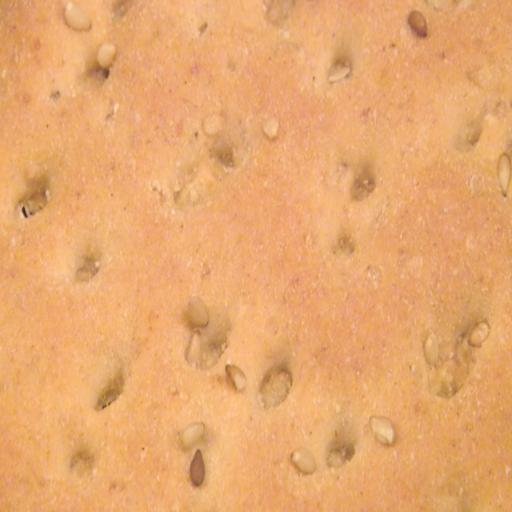} \\
\end{tabular}
}
\caption{Results of different approaches on $256\times 256$ to $512 \times 512$ texture synthesis. For SinGAN($^{*}$) and Non-stat.$^{(*)}$ results, the first two rows show the results when training with direct access to exact-size ground truth; the remaining 2 rows show the results without them accessing the ground truth.}
\label{fig:result512}
\end{figure*}

Table~\ref{tab:timing} shows the runtime and corresponding properties for all the methods. Compared with \textbf{Self-tuning}, our method is much faster. In contrast to \textbf{Non-stat.} and \textbf{SinGAN}, transposer (ours) generalizes better and hence does not require per image training. Comparing with \textbf{DeepTexture} and the style transfer method \textbf{WCT}, our method is still much faster without the need of iterative optimization or SVD decomposition. Even though \textbf{pix2pixHD} is faster than our method, it cannot perform proper texture synthesis as shown in Figures~\ref{fig:result256} and~\ref{fig:result512}, same as Texture Mixer~\cite{yu2019texture}. %mentioned in Section~\ref{sec:intro} and 

\textbf{Evaluation Metrics}. To the best of our knowledge, there is no standard metric to quantitatively evaluate texture synthesis results. 
%Yu et al.~\cite{yu2019texture} proposed some metrics to evaluate texture interpolation between two texture images which is not directly suitable for evaluating single texture image synthesis results. 
We use 3 groups of metrics (6 in total): 
\begin{enumerate}
% \vspace{-0.25cm}
\item \textbf{Existing metrics} include SSIM~\cite{wang2004image}, Learning Perceptual Image Patch Similarity (LPIPS)~\cite{zhang2018unreasonable} and Fr\'echet Inception Distance (FID)~\cite{heusel2017gans}. SSIM and LPIPS are evaluated using image pairs. FID measures the distribution distance between the generated image set and the ground truth image set in feature space.
% \vspace{-0.25cm}
\item \textbf{Crop-based metrics designed for texture synthesis evaluation} include crop-based LPIPS (c-LPIPS) and crop-based FID (c-FID). While the original LPIPS and FID are computed on full-size images, c-LPIPS and c-FID operate on crops of images. For c-FID,  we crop a set of images from the output image and crop the other set from the ground truth image, and then compute the FID between these two sets (we use a dimension of 64 for c-FID instead of the default 2048 due to a much smaller image set). For c-LPIPS, we compute the LPIPS between the input image and one of the 8 random crops from the output image, and average the scores among the 8 crops. 
% \vspace{-0.25cm}
\item \textbf{User Study}. Another way to measure the performances of different methods is by performing user study. We thus use Amazon Mechanical Turk (AMT) to evaluate the quality of synthesized textures. We perform AB tests where we provide the user the input texture image and two synthesized images from different methods. We then ask users to choose the one with better quality. Each image is viewed by $20$ workers, and the orders are randomized. The obtained preference scores (Pref.) are shown in Table~\ref{tab:eval200}, which indicate the portion of workers that prefer our result over the other method.
\end{enumerate}

\begin{figure*}
\vspace{-0.4cm}
\centering
\scalebox{0.6}{
\addtolength{\tabcolsep}{-4pt}
\begin{tabular}{cc|cc|cc|cc}
input & transposer(ours) & input & transposer(ours) & input & transposer(ours) \\
\includegraphics[width=0.07\textwidth]{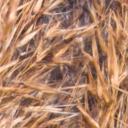}&
\includegraphics[width=0.28\textwidth]{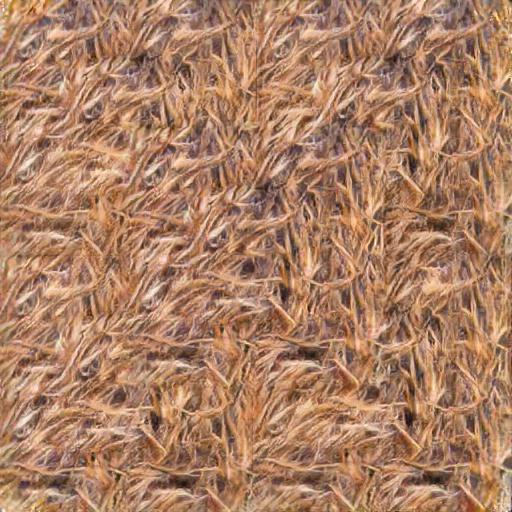}&
\includegraphics[width=0.07\textwidth]{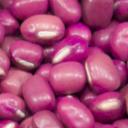}&
\includegraphics[width=0.28\textwidth]{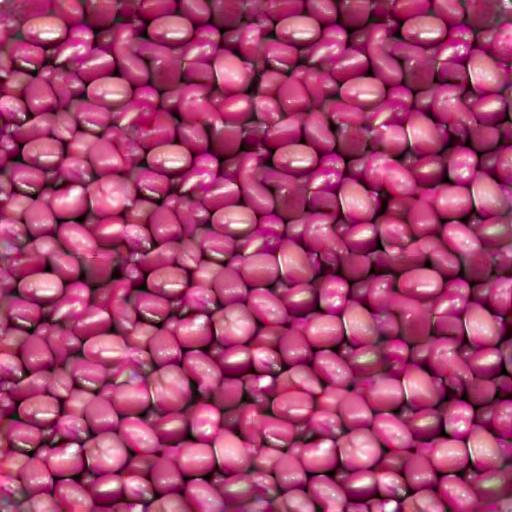}&
\includegraphics[width=0.07\textwidth]{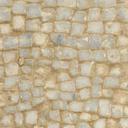}&
\includegraphics[width=0.28\textwidth]{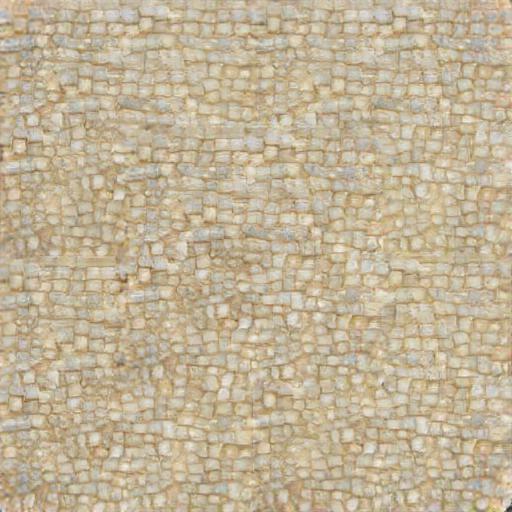} &
\includegraphics[width=0.07\textwidth]{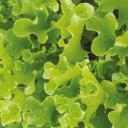}&
\includegraphics[width=0.28\textwidth]{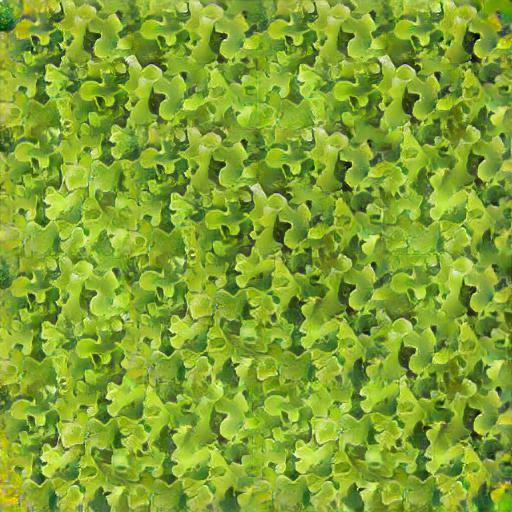} \\
\includegraphics[width=0.07\textwidth]{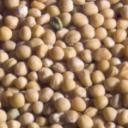}&
\includegraphics[width=0.28\textwidth]{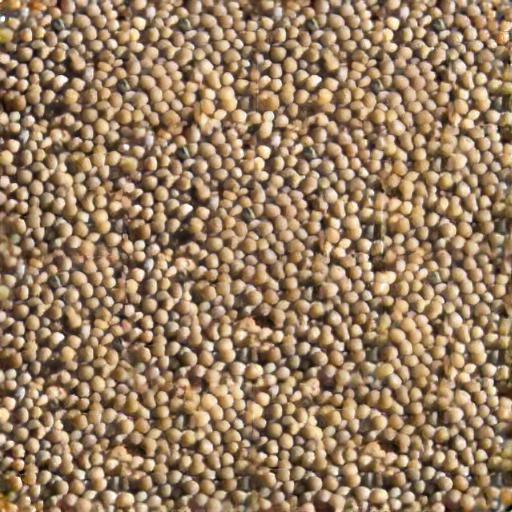} &
\includegraphics[width=0.07\textwidth]{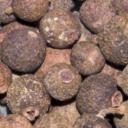}&
\includegraphics[width=0.28\textwidth]{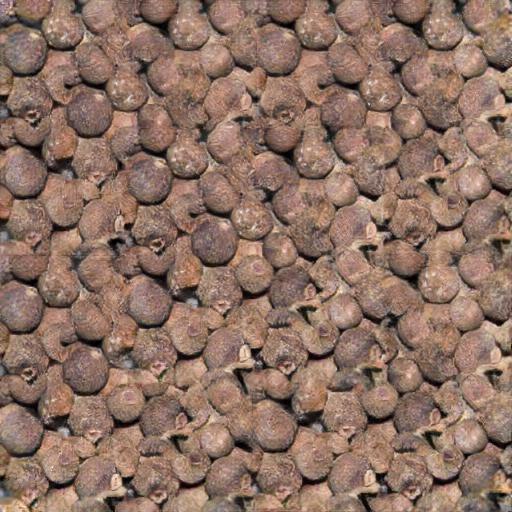} &

\includegraphics[width=0.07\textwidth]{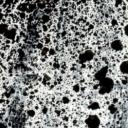}&
\includegraphics[width=0.28\textwidth]{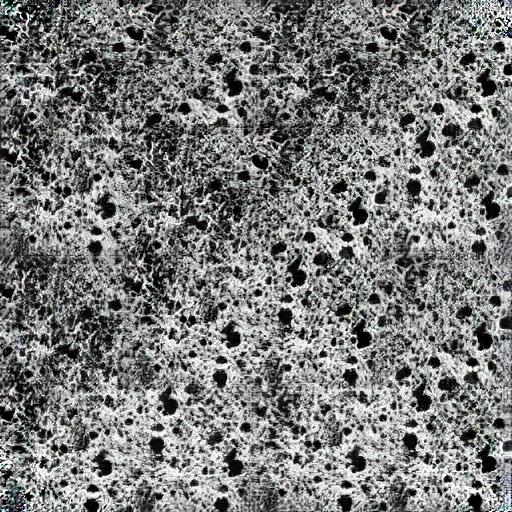} &
\includegraphics[width=0.08\textwidth]{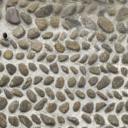} &
\includegraphics[width=0.28\textwidth]{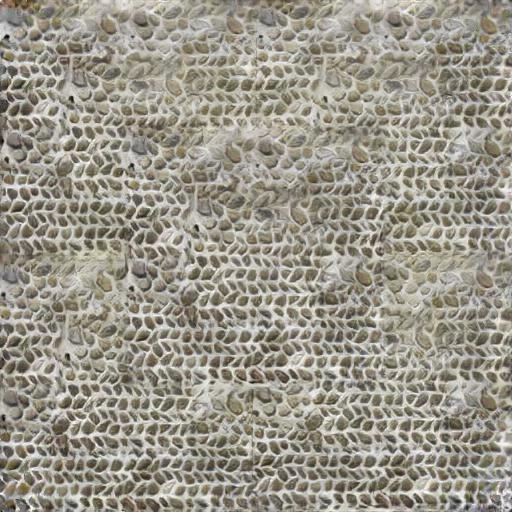} \\
\includegraphics[width=0.07\textwidth]{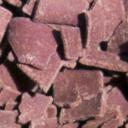}&
\includegraphics[width=0.28\textwidth]{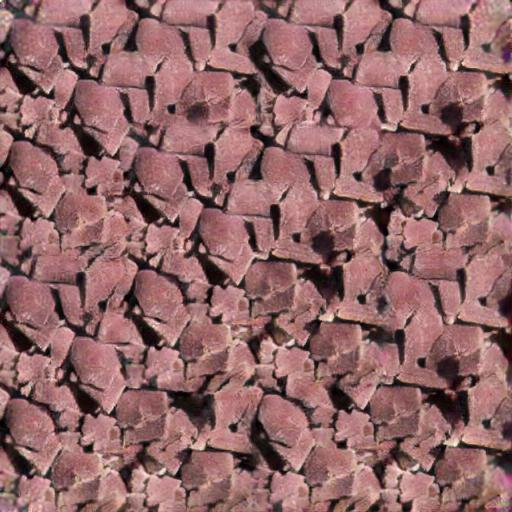}&
\includegraphics[width=0.07\textwidth]{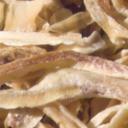}&
\includegraphics[width=0.28\textwidth]{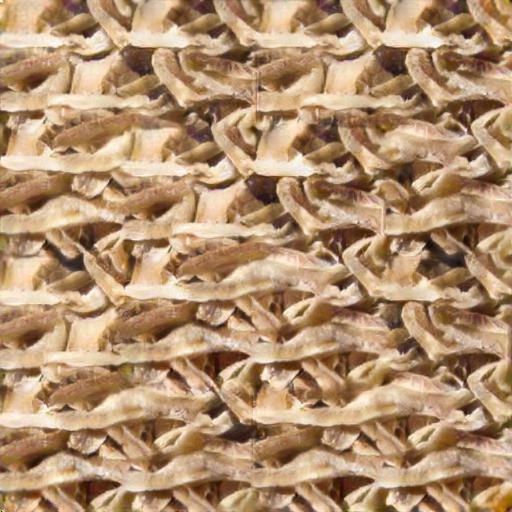} &
\includegraphics[width=0.07\textwidth]{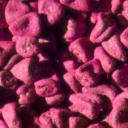}&
\includegraphics[width=0.28\textwidth]{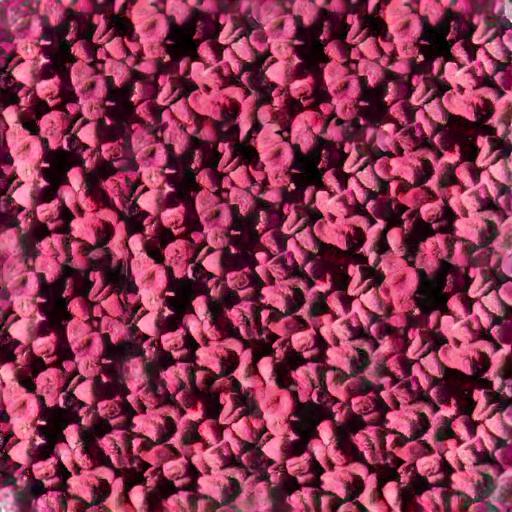} &

\includegraphics[width=0.07\textwidth]{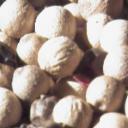}&
\includegraphics[width=0.28\textwidth]{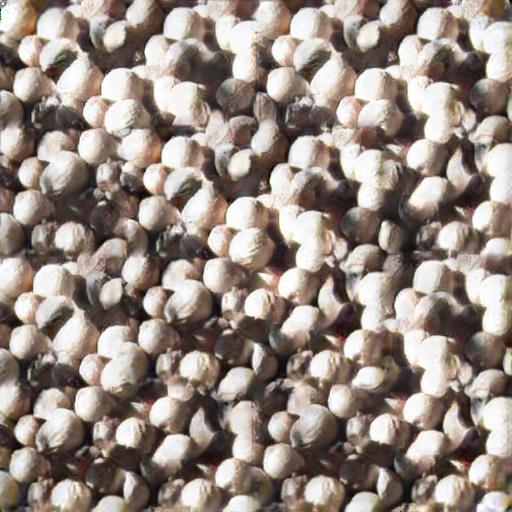} \\
\includegraphics[width=0.07\textwidth]{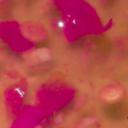}&
\includegraphics[width=0.28\textwidth]{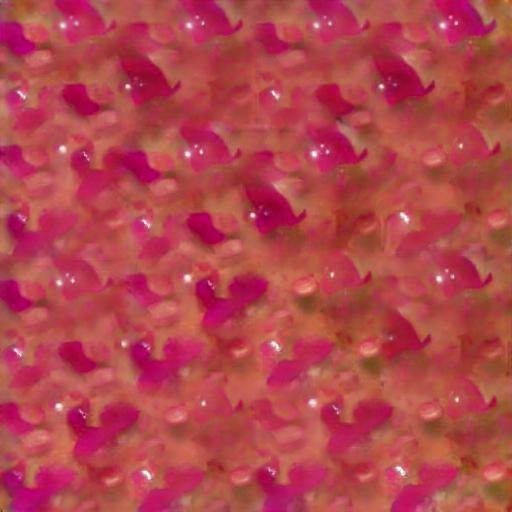} &

\includegraphics[width=0.07\textwidth]{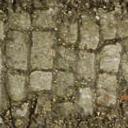}&
\includegraphics[width=0.28\textwidth]{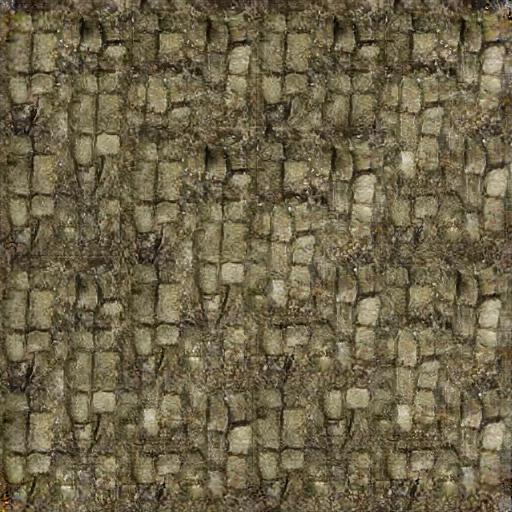} &
\includegraphics[width=0.07\textwidth]{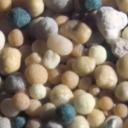}&
\includegraphics[width=0.28\textwidth]{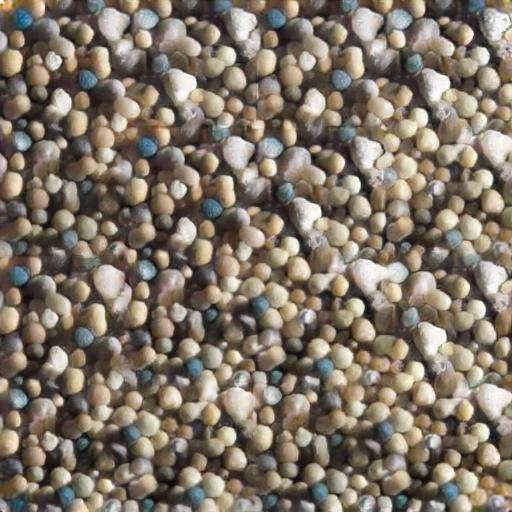} &
\includegraphics[width=0.07\textwidth]{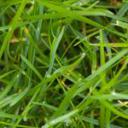}&
\includegraphics[width=0.28\textwidth]{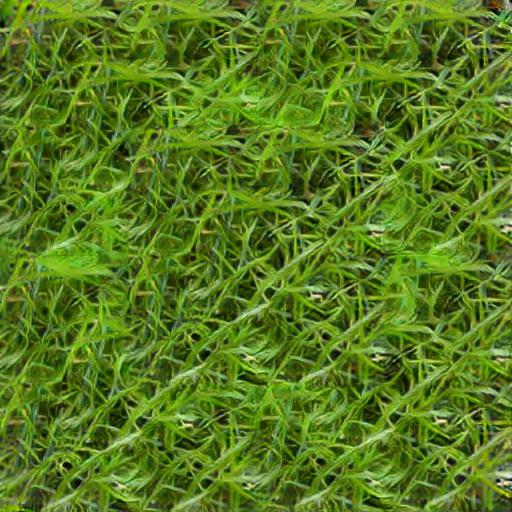} \\
% \includegraphics[width=0.07\textwidth]{Suppl128to512_More/FloorsPortuguese0062_350_input.jpg}&
% \includegraphics[width=0.28\textwidth]{Suppl128to512_More/FloorsPortuguese0062_350_output_output.jpg} 
% \\

\end{tabular}
}
\caption{More of our results on 4$\times$4 times larger synthesis. Zoom in for more details.}
\label{fig:more_4x4_result}
\vspace{-0.4cm}
\end{figure*}

\subsection{Comparison Results}
\subsubsection{Evaluating synthesis of size 128 to 256}

\begin{table}
    \centering
    \scalebox{0.92}{
    % \addtolength{\tabcolsep}{-2pt}
    \begin{tabular}{l|ccccc}
    \multicolumn{1}{c}{} & \multicolumn{1}{|c}{SSIM} & FID & c-FID & LPIPS & c-LPIPS \\
    \hline
    \hline
    Naive tiling & 0.311 & 23.868 & 0.5959 & 0.3470 & 0.2841\\
    Self-tuning & 0.3075 & 33.151 & 0.5118 & 0.3641 & 0.2970\\
    pix2pixHD & 0.318 & 26.800 & 0.5687 & 0.3425 & 0.2833 \\
    WCT & 0.280 & 57.630 & \textbf{0.4347} & 0.3775 & 0.3226 \\
    \textbf{transposer (ours)} & \textbf{0.386} & \textbf{21.615} & 0.4763 & \textbf{0.2709}  & \textbf{0.2653} \\
    \hline
    Ground Truth$^{*}$  & 1 & 0 & 0.1132 & 0 & 0.2670 \\
    \end{tabular}
    % \addtolength{\tabcolsep}{2pt}
    }
    \caption{Synthesis scores for different approaches averaged over 5,000 images.}
    \label{tab:eval5000}
    \vspace{-0.2cm}
\end{table}

We compare with Self-tuning, pix2pixHD and WCT on the whole test set of 5,000 images and show the quantitative comparisons in Table~\ref{tab:eval5000}. It is noticeable that our method outperforms Self-tuning and pix2pixHD for all the metrics. 
%Note that WCT$^{*}$ does not expand output size, and thus takes ground truth as input whereas other methods receive only a center crop of ground truth. We present non-expanding versions of \cite{zhou2018non,shaham2019singan} as well.
%\footnote{Method receives all of ground truth as input as opposed to a center-crop.\label{gt}}

Due to the fact that Non-stat., SinGAN and DeepTexture are too slow to evaluate on all 5,000 test images, we randomly sampled 200 from the 5,000 test images to evaluate them. The visual comparison is shown in Figure~\ref{fig:result256}. The numerical evaluation results are summarized in Table~\ref{tab:eval200}. As shown in 2nd-8th rows of Table~\ref{tab:eval200}, our method significantly out-performs all the methods which do not directly take ground truth as input. When compared  with Self-tuning, we achieve better LPIPS score (0.273 vs. 0.358), and 63$\%$ of people prefer the results generated by our method over the ones generated by Self-tuning. The remaining rows of Table~\ref{tab:eval200} also show that our method performs better than other size-increasing baselines (Non-stat.$^{*}$ and SinGAN$^{*}$) and performs better or similar to DeepTexture$^{*}$, which all take ground truth as input. For instance, 51$\%$ people prefer our results over the ground truth and 46$\%$ of people prefer our results over DeepTexture, which directly takes ground truth for its optimization. 

\begin{table}
    \centering
    \scalebox{0.9}{
    % \addtolength{\tabcolsep}{-4pt}
    \begin{tabular}{l|cccccc}
    \multicolumn{1}{c}{} & \multicolumn{1}{|c}{SSIM} & FID & c-FID & LPIPS & c-LPIPS & Pref. \\
    \hline
    \hline
    Naive tiling & 0.289 & 77.54 & 0.552 & 0.349 & 0.287 & - \\
    Self-tuning & 0.296 & 101.75 & 0.464 & 0.358 & 0.292 & 0.63 \\
    % Non-stationary & 227.607 & 0.3378 & 0.4803 & 0.3378 & 0.4203 \\
    % SinGAN & 164.399 & 0.3065 & 0.3733 & 0.3065 & 0.3039 \\
    Non-stat. &  0.321 & 143.31 & 2.728 & 0.3983 & 0.3436 & 0.92 \\
    SinGAN  & 0.337 & 212.30 & 1.375 & 0.3924 & 0.3245 & 0.81 \\
    pix2pixHD & 0.299 & 93.70 & 0.456 & 0.354 & 0.292 & 0.66 \\
    WCT & 0.280 & 126.10 & 0.401 & 0.375 & 0.300 & 0.67 \\
    Texture Mixer & 0.311 & 211.78 & 1.997 & 0.399 & 0.334 & 0.89 \\
    \textbf{transposer(ours)} & \textbf{0.437} & 74.35 & 0.366  & \textbf{0.273}& \textbf{0.272} & \\
    \hline
    Ground Truth$^{*}$ & 1 & 0 & 0.112 & 0 & 0.270 & 0.51 \\
    Non-stat.$^{*}$ &  0.767 & 73.72 & 2.149 & 0.1695 & 0.3276 & - \\
    SinGAN$^{*}$ & 0.492 & 88.14 & 1.137 & 0.2467 & 0.2939 & - \\
    DeepTexture$^{*}$ & 0.289 & \textbf{67.89} & \textbf{0.289} & 0.336 & 0.298 & 0.46 \\
    \end{tabular}
    }
    % \addtolength{\tabcolsep}{4pt}
    \caption{Synthesis scores for different approaches averaged over 200 images.}
    \label{tab:eval200}
    \vspace{-0.2cm}
\end{table}

\subsubsection{Evaluating synthesis of size 256 to 512}

We also evaluate on 256$\times$256 image to 512$\times$512 image synthesis using the same metrics. We show the quantitative results in the supplementary file. Visual comparisons can be found in Figure~\ref{fig:result512}.
% Table~\ref{tab:eval256to512} shows quantitative results. Both Table~\ref{tab:eval256to512} and Figure~\ref{fig:result512} 
It confirms that our approach produces superior results. For example, Self-tuning almost completely misses the holes in the 1st input texture image, and pix2pixHD simply enlarges the local contents instead of performing synthesis. In Figure~\ref{fig:more_4x4_result}, we show the 4$\times$4 times laerger texture synthesis results using our framework. This is done by running the transformer network twice with each performing 2$\times$2 times laerger synthesis.

%% file: sec_ablation.tex
\begin{table}
    \centering
    \scalebox{0.95}{
    \addtolength{\tabcolsep}{-2pt}
    \begin{tabular}{l|ccccc}
    \multicolumn{1}{c}{} & \multicolumn{1}{|c}{SSIM} & FID & c-FID & LPIPS & c-LPIPS \\
    \hline
    \textbf{Self-sim. Map (default)} & \textbf{0.437}  & \textbf{74.35} & \textbf{0.366}  & 0\textbf{0.273} & 0.272 \\
    \hline
    Learnable TransConv  & 0.3087  & 88.05  &  0.387 & 0.331  & 0.2797  \\
    Fixed Map  & 0.2966  & 97.79  & 0.383  & 0.3554  & 0.2848  \\     
    Random Map  & 0.2959  & 76.51  & 0.387 &  0.336 & \textbf{0.2645} \\
    \end{tabular}
    }
    \caption{Ablation study for transposed convolution operation and self-similarity map. For SSIM, the higher the better; for other metrics, the lower the better. The first row represents the transposer framework taking self-similarity map as inputs, the default setting in this paper. }
    \label{tab:eval_ablate}
    \vspace{-0.2cm}
\end{table}

\begin{figure*}
\centering
\scalebox{0.8}{
% \addtolength{\tabcolsep}{-4pt}   
% \begin{tabular}{c|cccc|c|cccc}
% \multicolumn{1}{c}{} & \multicolumn{4}{c}{} & \multicolumn{1}{c}{} & \multicolumn{4}{c}{} \\
\begin{tabular}{c|cccc}
\multicolumn{1}{c}{} & \multicolumn{4}{c}{}\\
Input & Ours & Learn. TransConv & Fixed Map & Random Map \\
\includegraphics[width=0.10\textwidth]{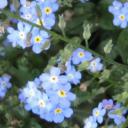} &
\includegraphics[width=0.20\textwidth]{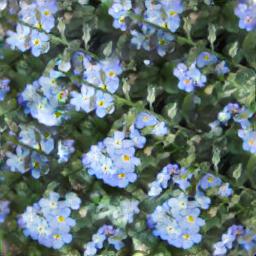} &
\includegraphics[width=0.20\textwidth]{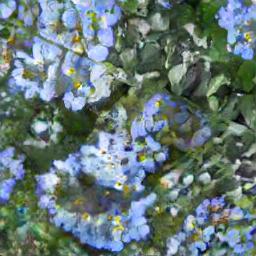} &
\includegraphics[width=0.20\textwidth]{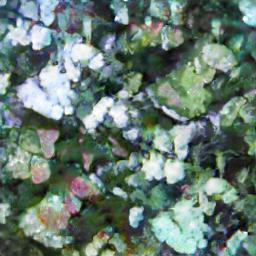} & 
\includegraphics[width=0.20\textwidth]{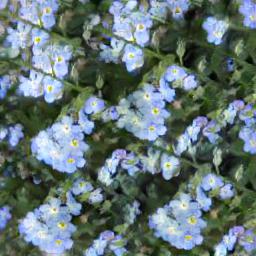} \\
\includegraphics[width=0.10\textwidth]{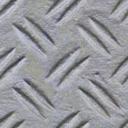} &
\includegraphics[width=0.20\textwidth]{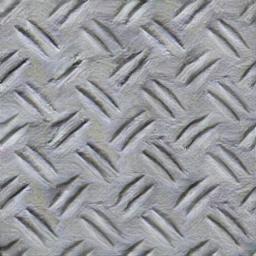} &
\includegraphics[width=0.20\textwidth]{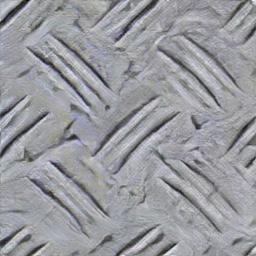} &
\includegraphics[width=0.20\textwidth]{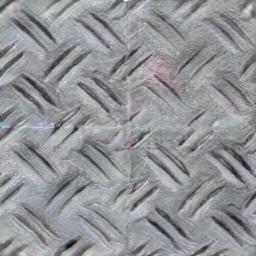} & 
\includegraphics[width=0.20\textwidth]{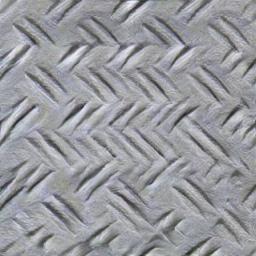} \\
\end{tabular}
}
\caption{Ablation study for using transposed convolution operations and self-similarity maps. It can be seen that without using them, the results become much worse.}
\label{fig:vis_ablate}
\end{figure*}

\begin{figure*}[!h]
\centering
\addtolength{\tabcolsep}{-3pt}
\small
\scalebox{0.96}{
\begin{tabular}{cccc}
\includegraphics[width=0.04\textwidth]{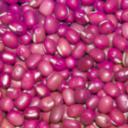} &
\includegraphics[trim=64 64 64 64,clip,width=0.45\textwidth]{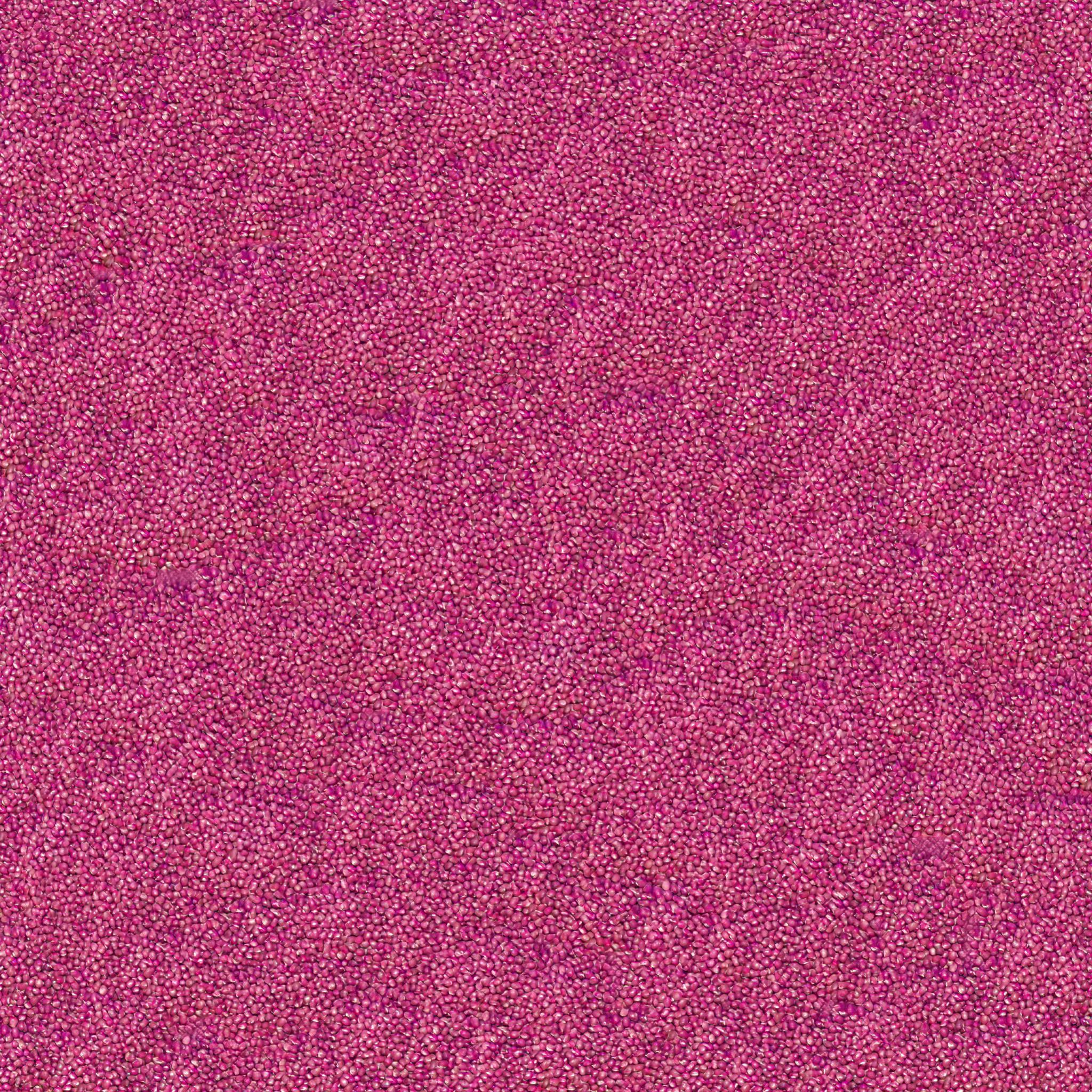} &
\includegraphics[width=0.04\textwidth]{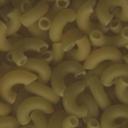} &
\includegraphics[trim=64 64 64 64,clip,width=0.45\textwidth]{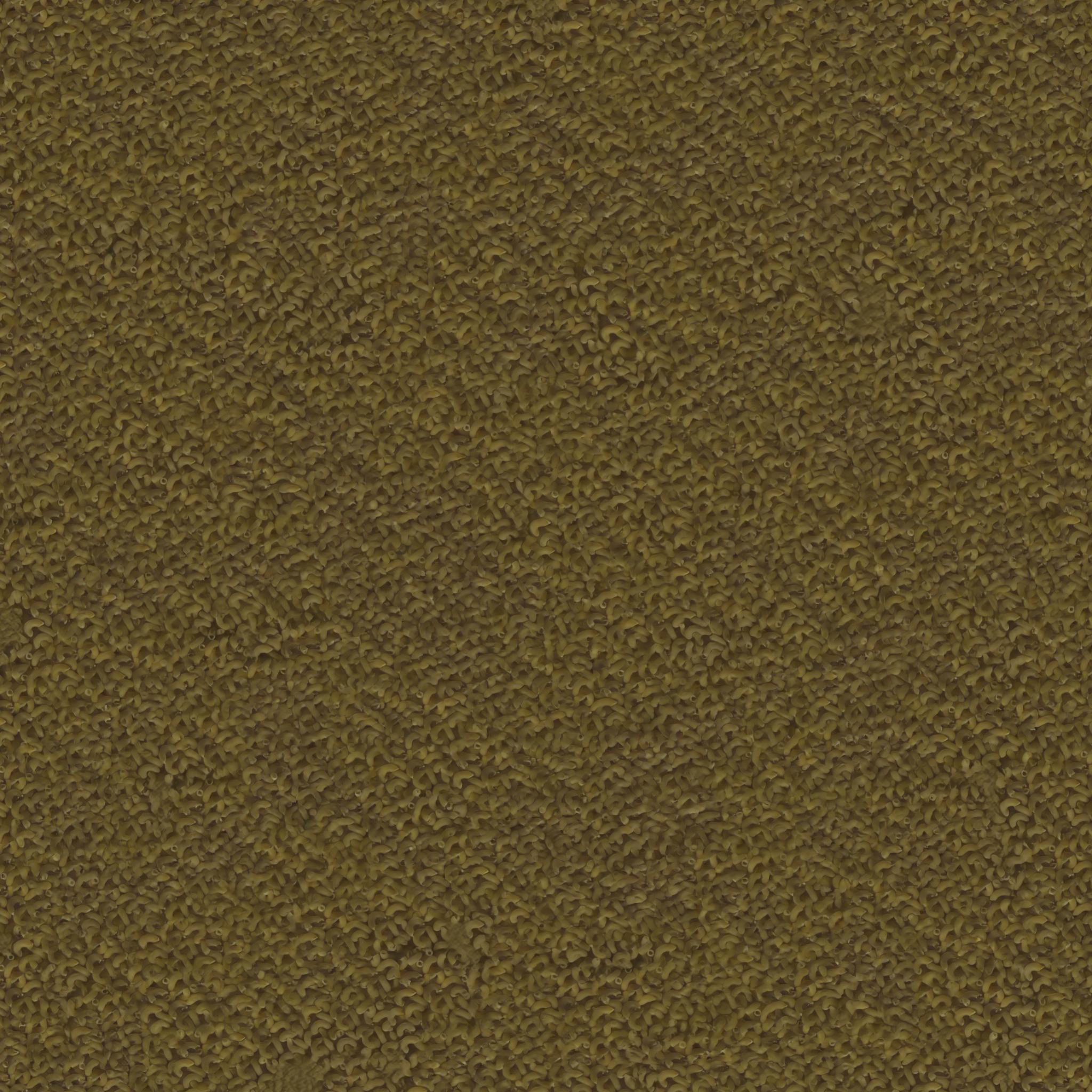} \\
\end{tabular}
}
\caption{Direct 2048$\times$2048 texture generation from 128$\times$128 input by sampling random noise maps. Zoom in for more details. Left small image is the input; right large image is the output.}
\label{fig:noise_big} 
\end{figure*}

\begin{figure}
\centering
\addtolength{\tabcolsep}{-3pt}
\small
\scalebox{0.9}{
\begin{tabular}{cccc}
input & result 1 & result 2 & result 3 \\
\includegraphics[width=0.07\textwidth]{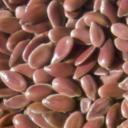} &
\includegraphics[width=0.14\textwidth]{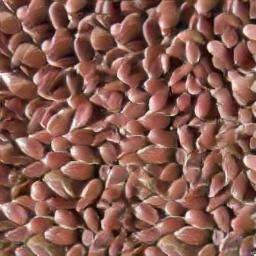} &
\includegraphics[width=0.14\textwidth]{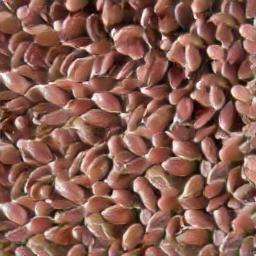} &
\includegraphics[width=0.14\textwidth]{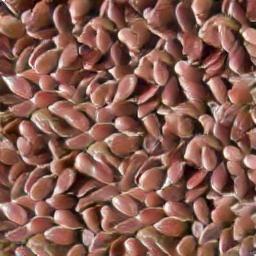} \\
\includegraphics[width=0.07\textwidth]{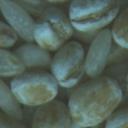} &
\includegraphics[width=0.14\textwidth]{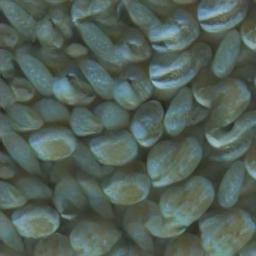} &
\includegraphics[width=0.14\textwidth]{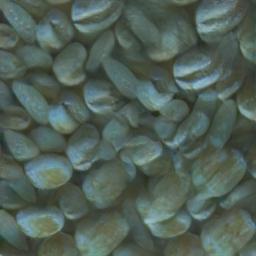} &
\includegraphics[width=0.14\textwidth]{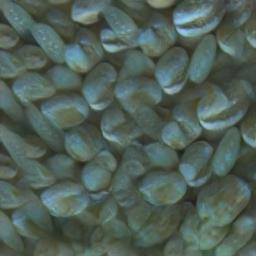} \\
\includegraphics[width=0.07\textwidth]{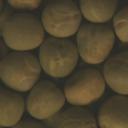} &
\includegraphics[width=0.14\textwidth]{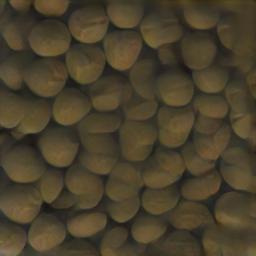} &
\includegraphics[width=0.14\textwidth]{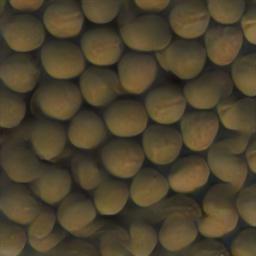} &
\includegraphics[width=0.14\textwidth]{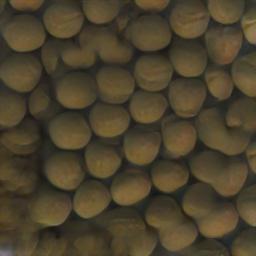} \\
\end{tabular}
}
\caption{Diverse outputs given different random noises as inputs.}
\label{fig:noise_diverse} 
\end{figure}

\subsection{Ablation Study and Random Noise as Input}

\subsubsection{Ablation study}

To understand the role of self-similarity map, we conduct three additional ablation study experiments: 1). \textbf{Learnable TransConv}: using the traditional transposed convolution layer with learnable parameters instead of directly using encoded feature as filters and its self-similarity map as input, while keeping other network parts and training strategies unchanged; 2). \textbf{Fixed Map}: using fixed sampled maps instead of self-similarity maps; 3). \textbf{Random Map}: using randomly sampled maps instead of self-similarity maps. As shown in Figure~\ref{fig:framework}, we have 3 different scales' features, for running \textbf{Fixed map} and \textbf{Random map}, we sample the map for the smallest scale's feature and then bilinear upsampling it for the other two scales. Table~\ref{tab:eval_ablate} and Figure~\ref{fig:vis_ablate} show the quantitative and qualitative results, respectively. These 3 settings are compared with the default transformer setting, using self-similarity map as transposed convolution input. It can be seen that \textbf{Learnable TransConv} with the traditional learnable transposed convolution layer will simply enlarge the input rather than perform reasonable synthesis, similar to pix2pixHD. This confirms our hypothesis that conventional CNN designs with traditional (de)convolution layers and up/down-sampling layers cannot capture the long-term structural dependency required by texture synthesis. \textbf{Fixed map} can't produce faithful results. On the other hand, using random noise map as transposed convolution input has both advantages and disadvantages, as discussed below.

\subsubsection{Trade-off between self-similarity map and random noise map}

In the last column of Figure~\ref{fig:vis_ablate}, the 1st row shows that sampling a random noise map at test time can successfully generate diverse results. However, note that the self-similarity map is critical in identifying the structural patterns and preserving them in the output. In the 2nd row of Figure~\ref{fig:vis_ablate}, the result of using self-similarity maps successfully preserved the regular structures, while using random noise maps failed. We believe that in practice, there is a trade-off between preserving the structure and generating variety. For input texture images with regular structural patterns, self-similarity map provides better guidance for the transposed convolution operation to preserve these structural patterns. On the other hand, using random noise map as inputs can generate diverse outputs by sampling different noise maps, shown in Figure~\ref{fig:noise_diverse} and it is also possible to directly generate arbitrary large texture outputs by sampling larger noise map, shown in Figure~\ref{fig:noise_big} while using self-similarity map can only do smaller than 3$\times$3 times larger synthesis, limited by the size of self-similarity map.

%% file: sec_discuss.tex
\section{Conclusion \& Discussion}

\begin{figure}{r} %{0.5\textwidth}
  \begin{center}
    \includegraphics[width=0.08\textwidth]{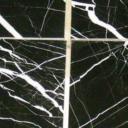}
    \includegraphics[width=0.16\textwidth]{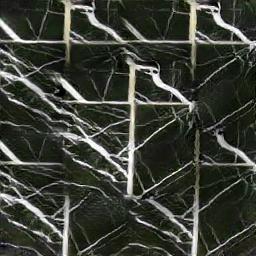}
    \includegraphics[width=0.16\textwidth]{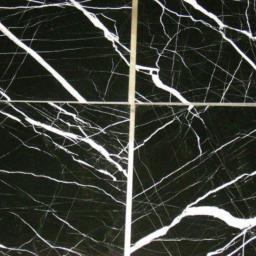}
  \end{center}
  \caption{Failure case for our method. From left to right: input, our synthesis result and the ground truth.}
  \label{fig:failure}
\end{figure}    

In this paper, we present a new deep learning based texture synthesis framework built based on transposed convolution operations. In our framework, the transposed convolution filter is the encoded features of the input texture image, and the input to the transposed convolution is the self-similarity map computed on the corresponding encoded features. Quantitative comparisons based on existing metrics, our specifically designed metrics for texture synthesis, and user study results all show that our method significantly outperforms existing methods, while our method also being much faster. Self-similarity map helps preserve the structure better while random noise map allows to generate diverse results. Some further research could also be providing more control-able flexibility by combining both self-similarity map and random noise map as inputs. One limitation of our method is that it fails to handle sparse thin structures like shown in Figure~\ref{fig:failure} and highly non-stationary inputs~\cite{zhou2018non}. As some highly non-stationary textures mainly emphasize the effect on some specific direction, one possible solution to deal with them may be emphasizing the similarity score on specific directions while suppressing it on other directions to capture directional effects, and/or using cropped, resized or rotated feature maps as transposed convolution filters to capture the effects of textons repeating with various forms. We leave these as future research exploration. While existing deep learning-based image synthesis methods mostly focus on taking the inputs from other modalities like semantic maps or edge maps, we believe our method will also stimulate more deep learning researches for exemplar-based synthesis.